\documentclass[journal]{IEEEtran}
\usepackage{amsfonts}
\usepackage{float}
\usepackage[comma,numbers,square,sort&compress]{natbib}
\usepackage{epstopdf}
\usepackage{subfigure}
\usepackage{graphicx}
\usepackage{amsthm}
\usepackage[top=2cm, bottom=2cm, left=2cm, right=2cm]{geometry}
\usepackage{algorithm}
\usepackage{algorithmicx}
\usepackage{algpseudocode}
\usepackage{amsmath}
\usepackage{multirow}
\usepackage{booktabs}
\usepackage{threeparttable}
\newcommand{\RNum}[1]{\uppercase\expandafter{\romannumeral #1\relax}}

\usepackage{fancyhdr}

\begin{document}

\title{A Novel Learning-based Global Path Planning Algorithm for Planetary Rovers}

\author{Jiang~Zhang,
        Yuanqing~Xia,
        Ganghui~Shen
\thanks{Jiang Zhang, Yuanqing Xia and Ganghui Shen are with the School of Automation, Beijing Institute of Technology, Beijing 100081, China. Email: bitzj2015@outlook.com (Zhang), xia\_yuanqing@bit.edu.cn (Xia), hxyzsgh@gmail.com (Shen).}}

\maketitle
\thispagestyle{fancy}
\fancyhead{} 
\lhead{} 
\lfoot{\begin{small}
\end{small}}
\cfoot{} 
\rfoot{}

\begin{abstract}
Autonomous path planning algorithms are significant to planetary exploration rovers, since relying on commands from Earth will heavily reduce their efficiency of executing exploration missions. This paper proposes a novel learning-based algorithm to deal with global path planning problem for planetary exploration rovers. Specifically, a novel deep convolutional neural network with double branches (DB-CNN) is designed and trained, which can plan path directly from orbital images of planetary surfaces without implementing environment mapping. Moreover, the planning procedure requires no prior knowledge about planetary surface terrains. Finally, experimental results demonstrate that DB-CNN achieves better performance on global path planning and faster convergence during training compared with the existing Value Iteration Network (VIN).
\end{abstract}


\section{Introduction}
During planetary exploration missions, rovers are required to explore diverse targets of interest after successful landing. Since the surfaces of planets (e.g. Mars and Moon) are covered with dangerous areas (e.g. rocks, steep slope, and craters) and the power supplied for rovers are limited, it is important for planetary rovers to find collision-free and energy-efficient paths to destination \cite{ref1}. Moreover, the uncertain planetary environments and the unavoidable communication delays between Earth and other planets make it impractical to provide real-time decision and control for rovers from Earth. This means that the design of autonomous path planning algorithms is indispensable for planetary rovers. 

The planetary path planning problem can be classified into two types, namely global path planning and local path planning. For global path planning, the whole trajectories from rovers' start positions to their targets are required to be determined from planetary surface images captured by orbit satellites. It can be fullfilled offline since global environments are totally observable. For local path planning, the partial trajectories from rovers' current positions to their ends of sight need to be planned from their observations of local environments. It is commonly executed online since only local environments are observable. This paper concentrates on dealing with global path planning problem for planetary rovers.
\begin{figure}[H]
\centering
\includegraphics[width=1\linewidth]{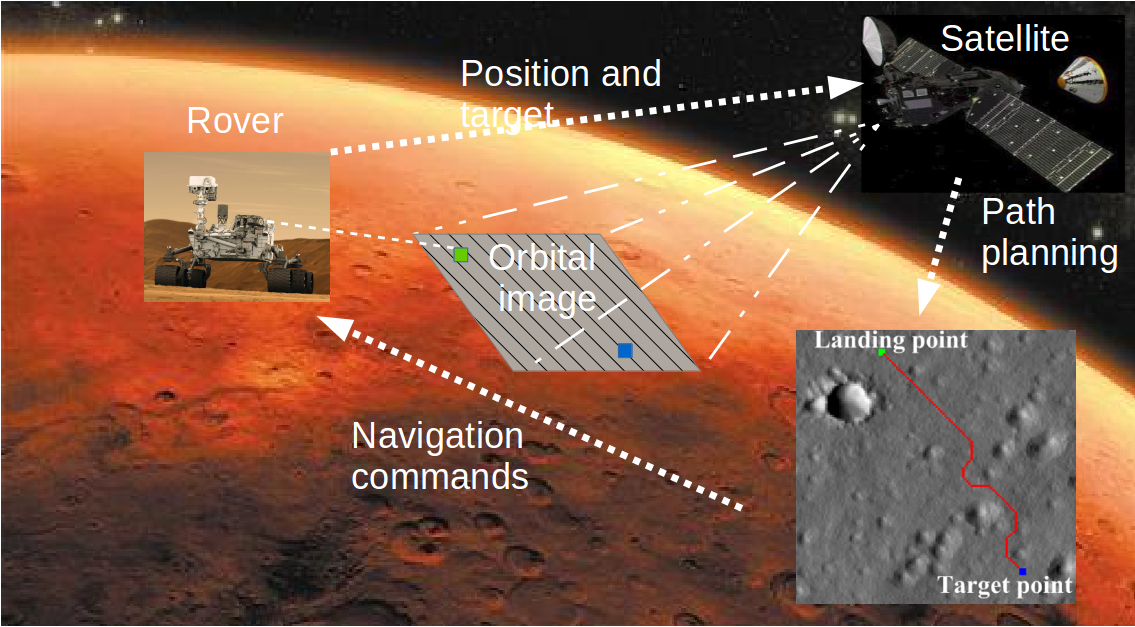}
\caption{Autonomous global path planning.}\label{Fig.1.1}
\end{figure}
Typically, the initial stage of implementing global path planning algorithms is mapping the real-world environment \cite{ref2}. More precisely,  observations of global environments are commonly transformed into  configuration space (C-space), visibility graph, Voronoi diagram or grid maps \cite{ref3,ref4}. Then, global path planning algorithms can be applied. In \cite{ref5,ref6}, classical shortest path search methods such as Dijkstra algorithm and Floyd algorithm were firstly employed to deal with global path planning problem. However, since global path planning with multiple obstacles is non-deterministic polynomial time hard (NP-hard) \cite{ref2}, it is time-consuming to find the shortest path through traversal search. Therefore, heuristic and evolutionary algorithms were adopted to address global path planning efficiently. 
In \cite{ref7,ref8}, heuristic search algorithms such as $A^*$ and $D^*$ were applied to achieve efficient path planning for mobile robots successfully. Then, inspired by natural and biological intelligence, evolutionary algorithms such as genetic algorithm \cite{ref9}, particle swarm optimization \cite{ref10}, ant colony algorithms \cite{ref11}, and neural network algorithms \cite{ref12} were extended into global path planning problems for planetary rovers. It is noteworthy that these algorithms cannot work without environment mapping, for which humans' prior knowledge about planetary environments are necessary.

In order to achieve autonomous path planning directly from orbital images, some algorithms have to firstly represent and learn deep features of orbital images such as the shape and location of obstacles. Then, according to these deep feature, the optimal path can be determined. In recent years, Deep Convolutional Neural Networks (DCNNs) have received wide attention in computer vision field for their superior feature representation and learning capability \cite{ref13}.
Inspired by the state-of-the-art performance of DCNNs in visual feature representation and learning, learning to plan directly from original images have been researched. Since global path planning is a sequential decision making process, one proven techique is formulating it as a Markov Decision Process (MDP) and finding the optimal path planning policy through value function estimation. In \cite{ref14}, a novel DCNN arthitecture---Value Iteration Network (VIN) was proposed to effectively estimate value functions in MDP. Then, the goal of planning path directly from Martian orbital images was achieved. Based on the work of VIN, Memory Augmented Control Network and Neural Map were proposed to find the optimal path for rovers in partially observable environment in \cite{ref15} and \cite{ref16} respectively. Further, in order to plan path for rovers under dynamic environments, Value Propagation Network \cite{ref17} was designed. However, all these networks contain the value iteration module in VIN, which has low training and planning efficiency since it requires multiple times of iteration inside the network for value function estimation.

Therefore, in this paper, we design a novel DCNN architecture with double branches  and non-iteration sturcture (DB-CNN) for value function estimation, which can achieve global path planning with higher efficiency and precision. The main contributions of this paper are summarized as follows:
\begin{itemize}
\item A novel DCNN architecture with double branches (DB-CNN) is designed to achieve autonomous global path planning direcly from planetary orbital images.
\item We present the global path planning algorithm based on DB-CNN and illstruate its merits over traditional global path planning methods.   
\item Compared with the state-of-the art architecture (VIN), DB-CNN achieves better performance and faster convergence on planetary global path planning tasks.
\item Experimental analysis demonstrates that DB-CNN has more efficient learning structure and the training time is alrgely reduced by compared with VIN.
\end{itemize}

The rest paper is organized as follows. Section \RNum{2} provides preliminaries of this paper. Section \RNum{3} describes the proposed DB-CNN for global path planning of planetary rovers. Experimental results and analysis are illustrated in Section \RNum{4}, followed by discussion and conclusions in Section \RNum{5}.
\section{Preliminaries}
\subsection{Markov Decision Process}
\begin{figure}[H]
\centering
\includegraphics[width=1\linewidth]{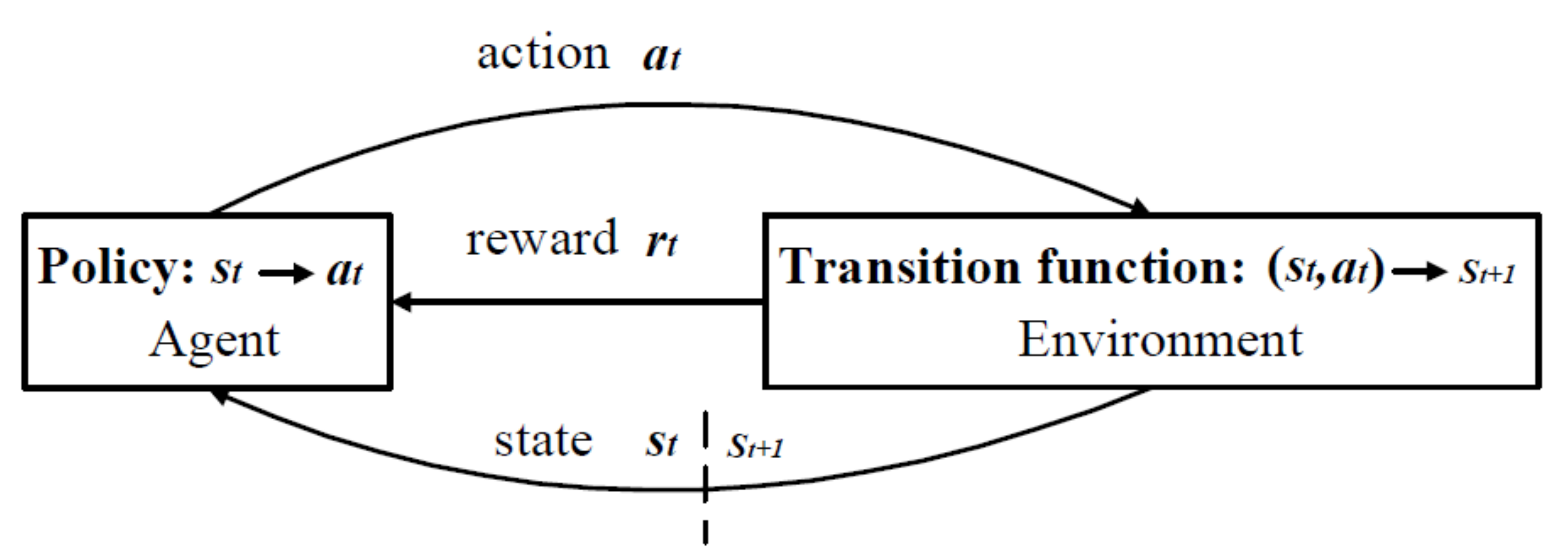}
\caption{Markov decision process.}\label{Fig.2.1}
\end{figure}

A standard MDP for sequential decision making is composed of action space $\mathcal{A}$, state space $\mathcal{S}$, reward fucntion $R:\mathcal{S}\times \mathcal{A}\rightarrow \mathbb{R}$, transition probability distribution $P:\mathcal{S}\times \mathcal{A} \times \mathcal{S}\rightarrow \mathbb{R}$ and discounted factor $\gamma$, where the policy is denoted by $\pi: \mathcal{S}\times \mathcal{A}\rightarrow \mathbb{R}$. At time step $t$, the agent can observe its state $s_t$ from environment and then choose its action $a_t$ satisfying $a_t \sim \pi(a|s_t), a \in \mathcal{A}$ (or $a_t = \pi(s_t)$ if the policy is deterministic). After that, its state will transit into $s_{t+1}$ and the agent will then receive reward $r_t=R(s_t,a_t)$ from environment, where $s_{t+1}$ satisfies $s_{t+1}\sim P(s|s_t,a_t), s \in \mathcal{S}$ (or $s_{t+1}=P(s_t,a_t)$ if the state trasition process is deterministic). The whole process is shown in Fig.~\ref{Fig.2.1}.

Furthermore, denote the discount factor of reward by $\gamma \in [0,1]$. The optimal policy is defined as
\begin{equation}
\label{eq2.1}
\pi^{*} = \arg\max_{\pi} E_{s_0}\Big[\sum_{i=0}^{+\infty}\gamma^{i}r_i|\pi,P\Big],
\end{equation}


To measure the expected accumulative reward of $s_t$ and $(s_t,a_t)$, the state value function and the action value function are defined as 
\begin{equation}
\label{eq2.2}
V(s_t)=E\Big[\sum_{i=t}^{+\infty}\gamma^{i-t}r(s_{i},a_{i})|\pi,P,s_t\Big],
\end{equation}
\begin{equation}
\label{eq2.3}
\begin{aligned}
Q(s_t,a_t)&=E\Big[\sum_{i=t}^{+\infty}\gamma^{i-t}r(s_{i},a_{i})|\pi,P,s_t,a_t\Big]\\
&=r_{t} + E\Big[V(s_{t+1})|\pi,P,s_{t+1}\Big].
\end{aligned}
\end{equation}

By substituting Eq.~(\ref{eq2.2}) into Eq.~(\ref{eq2.1}), the optimal policy is derived as
\begin{equation}
\label{eq2.4}
\pi^{*} = \arg\max_{\pi} E_{s_0}[V(s_0)]=\arg\max_{\pi} E_{s_0,a_0}[Q(s_0,a_0)].
\end{equation}

However, since both state value function and action value function are unknown, it is impossible to determine $\pi^{*}$ through Eq.~(\ref{eq2.4}) directly. Therefore, value functions of MDP have to be estimated precisely so that the optimal policy can be found.

\subsection{Value Function Estimation}
Value iteration is an typical method for value function estimation and then addressing MDP problem \cite{ref14}. Denote the estimated state value function at step $k$ by $V_{k}(s)$, and the estimated action value function for each state at step $k$ by $Q_{k}(s,a)$. $\pi_{k}$ is utilized to represent the deterministic policy at step $k$. Then, the value iteration process can be expressed as 
\begin{equation}
\label{eq2.5}
\pi_{k}(s_i) = \arg \max_{a_i}Q_{k}(s_i,a_i) \quad (i=0,1,\cdots),
\end{equation}
\begin{equation}
\begin{aligned}
\label{eq2.6}
V_{k+1}(s_i) &= Q_{k+1}(s_i,\pi_{k}(s_i)) \\
&= r_i +  E_{s_{i+1}}[V_{k}(s_{i+1})] \quad (i=0,1,\cdots).
\end{aligned}
\end{equation}

\begin{figure}[H]
\centering
\includegraphics[width=1\linewidth]{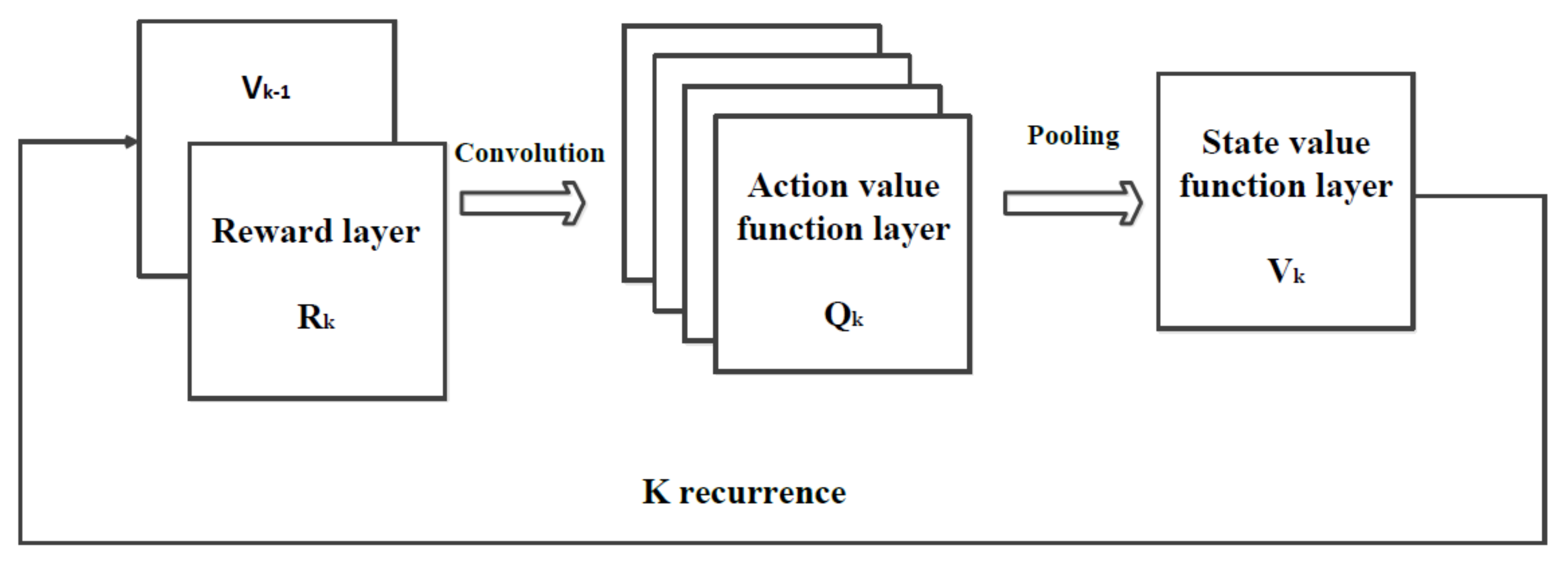}
\caption{Value iteration module.}\label{Fig2.2}
\end{figure}

However, since it is difficult to determine the explicit representation of $\pi_{k}$, $Q_{k}$ and $V_{k}$ (especially when the dimension of $s_t$ is high), VIN is designed to approximate this process successfully, which consists of Value Iteration Module. As illustrated in Fig.~\ref{Fig2.2}, the value function layer $V_k$ is stacked with the reward layer $R_k$ and then filtered by a convolutional layer and a max-pooling layer recurrently. Furthermore, through VIN, global information including orbital images and target position can be conveyed to each state in the final value function layer. Experiments demonstrate that this architecture performs well in learning to plan tasks. However, it takes lots of time to train such a recurrent convolutional neural network especially when the value of iteration time ($K$ in Fig.~\ref{Fig2.2}) becomes large. Therefore, replacing Value Iteration Module with a more efficient architecture without losing its excellent global path planning performance becomes the focus of this paper.

\subsection{Methods for Value Function Estimation}
Generally, there exist two learning-based methods for value function estimation---reinforcement learning \cite{ref18} and imitation learning \cite{ref19}. In reinforcement learning, no prior knowledge is required and the agent can find the optimal policy in complex environment by trial and error \cite{ref20}. However, the training process of reinforcement learning is computationally inefficient. In Imitation learning, when the expert dataset is given $\{(s_i, y_i)\}_{i=1}^{i=N}$, the training process transforms into supervised learning with higher data-efficiency and fitting accuracy.

Considering that the expert dataset $\{(s_i, y_i)\}_{i=1}^{i=N}$ for global path planning is available ($y_i \in \{0,1,\cdots,7\}$ is the optimal action at state $s_t$ and $N$ is the number of samples), in this paper, imitation learning method is applied to find the optimal navigation policy.
\section{Model Description}
\subsection{Global Path Planning Model}
In this subsection, we formulate the global path planning problem of planetary rovers into a MDP defined as $\mathcal{M}=<\mathcal{S},\mathcal{A},P,R>$. 
\subsubsection{State Space $\mathcal{S}$}
The state space of $\mathcal{M}$ is denoted as $\mathcal{S} =\{\mathcal{I},\mathcal{G},\mathcal{X}\}$, consisting of $\mathcal{I}=\{I_t\}$, $\mathcal{G}=\{(g_{t}^1,g_{t}^2)\}$ and $\mathcal{X}=\{(x_{t}^1,x_{t}^2)\}$. More precisely, $I_t\in\mathbb{R}^{H\times W\times C}$ is the planetary orbital image at time step $t$ with height $H$, width $W$, and $C$ channels, $(g_{t}^1,g_{t}^2) \in \{0,1,\cdots,H-1\}\times \{0,1,\cdots,W-1\}$ is the target position for planetary rover at time step $t$, and $(x_{t}^1,x_{t}^2)\in \{0,1,\cdots,H-1\}\times \{0,1,\cdots,W-1\}$ is rover's location at time step $t$.
\subsubsection{Action Space $\mathcal{A}$} 
The action space of $\mathcal{M}$ is denoted as $\mathcal{A} = \{a_t|a_t \in \{0,1,\cdots,7\}\}$, representing eight potential moving direction of planetary rover (0: east, 1: south, 2: west, 3: north, 4: southeast, 5: northeast, 6: southwest, 7:northwest).
\subsubsection{State Transition Function $P$}
Since state transition process in this MDP is deterministic, it is defined as $P: \mathcal{S}\times\mathcal{A}\rightarrow\mathcal{S}$. After taking action $a_t$, the state $s_t$ will transit into $s_{t+1}$. Notably, for given exploration mission, the planetary orbital image $I_t$ and the target position for planetary rover $(g_{t}^1,g_{t}^2)$ in state $s_t$ are constant during each path planning step $t$ while the rover's position $(x_{t}^1,x_{t}^2)$ in state $s_t$ will change at each step. 
\subsubsection{Reward Function $R$}
If the rover reaches the target point precisely at time step $t+1$ after taking action $a_t$, it will obtain a positive reward $r_t=\phi_1$ ($\phi_1 > 0$). Otherwise, it will get a negative reward $r_t=\phi_2$ ($\phi_2 < 0$). Therefore, the optimal path from start position to target position will have the maximal accumulative rewards.
\subsubsection{Problem Formulation}
Denote the DCNN designed for value function estimation as $F_{\alpha}: s_t \rightarrow softmax[\hat Q(s_t,a=0), \cdots, \hat Q(s_t,a=7)]^{T}$, where $\alpha$ represents the parameter of this DCNN and $\hat Q(s,a)$ is the estimated value of $Q(s,a)$. Then, the policy for global path planning is derived as 
\begin{equation}
\label{eq2.8}
a_t=\pi(s_t)=\arg \max_{a}\hat Q(s_t,a).
\end{equation}

Given the expert dataset $\{(s_i, y_i)\}_{i=1}^{i=N}$ for global path planning, we can view this DCNN as a classifier with 8 classes and define the training loss in cross entropy form with $L_2$ norm \cite{ref21} as follows
\begin{equation}
\begin{aligned}
\label{eq2.7}
L(\alpha) = -\frac{1}{N}\sum_{i=1}^{N}Y_i {\rm log}( F_{\alpha}(s_i)) + \lambda ||\alpha||_2,
\end{aligned}
\end{equation}
where $N$ is the number of training samples, $Y_i$ is the one-hot vector form \cite{ref22} of $y_i$ and $\lambda$ is the hyperparameter adjusting the effect of $L_2$ norm on the loss function.

By minimizing the loss function $L(\alpha)$, the optimal parameter of the DCNN can be determined as follows
\begin{equation}
\begin{aligned}
\label{eq2.10}
\alpha^* = \arg \min_{\alpha}L(\alpha).
\end{aligned}
\end{equation}

Therefore, the global path planning problem is formulated as designing and training a DCNN for value function estimation, which best fits the given expert dataset.

\begin{figure*}
\centering
\includegraphics[width=1\linewidth]{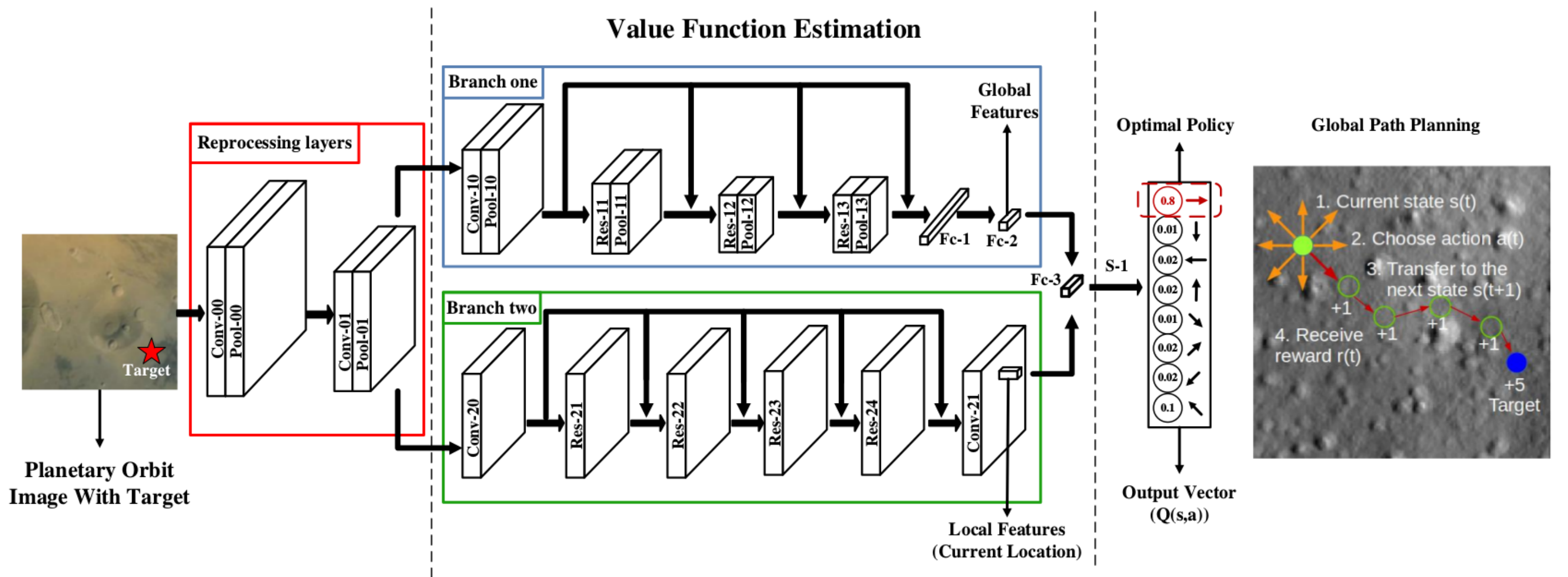}
\caption{DB-CNN for global path planning.}
\label{Fig.3.1}
\end{figure*}
\subsection{Proposed DB-CNN for Value Function Estimation}
In this subsection, we propose a novel deep neural network architecture for value funciton estimation---DB-CNN, which is composed of reprocessing layers, branch one for global feature representation, and branch two for local feature representation.
\subsubsection{Reprocessing Layers}
The reprocessing layers comprise of two convolutional layers (Conv-00, Conv-01), each of which is followed by one max-pooling layer (Pool-00, Pool-01). The aim of reprocessing layers is to filter out noise and compress the original orbit image $I_t$ into feature map $I_t^{'} \in \mathbb{R}^{H^{'}\times W^{'}\times C^{'}}(H = l_1H^{'}, W=l_2W^{'})$. After that, global path planning becomes $area~by~area$ with size $l_1\times l_2$ instead of $pixel~by~pixel$, the efficiency of which is improved.
\subsubsection{Branch One}
\begin{figure}[H]
\centering
\includegraphics[width=1\linewidth]{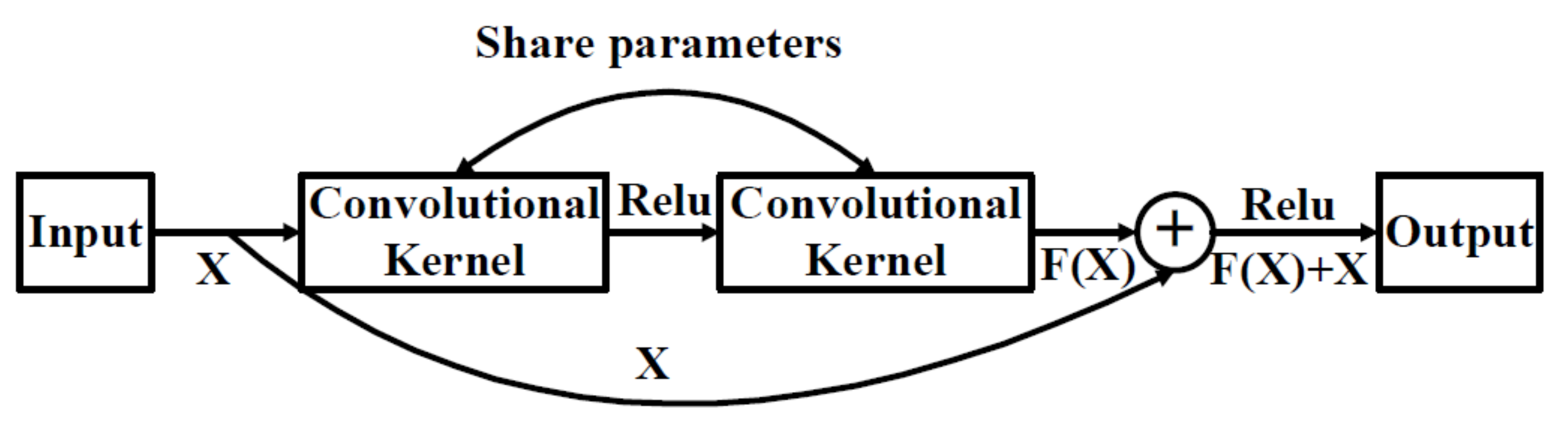}
\caption{Residual convolutional layer.}\label{Fig.3.3}
\end{figure}

Branch one consists of one convolutional layer (Conv-10), three residual convolutional layers (Res-11, Res-12, Res-13), four max-pooling layers (Pool-10, Pool-11, Pool-12, Pool-13) and two fully connected layers (Fc-1, Fc-2). Notably, residual convolutional layer (Fig.~\ref{Fig.3.3}) is one kind of convolutional layer proposed in \cite{ref23}, which not only increases the training accuracy of convolutional neural networks with deep feature representations, but also makes them generalize well to testing data. Considering that DB-CNN is required to represent deep features of orbital images and achieves high-precision under unknown environments (testing images), residual convolutional layers are embedded in DB-CNN. We denote the deep feature extracted from feature map $I_t^{'}$ by this branch as $f_1\in\mathbb{R}^{D}$ ($D$ is dimension of feature vector $f_1$). $f_1$ can be viewed as a global guidance to planetary rover, which represents global features related to all pixels in orbital image $I_t$ and target position $(g_{1t},g_{2t})$.

\subsubsection{Branch Two} 
Branch two is composed of two convolutional layers (Conv-20, Conv-21) and four residual convolutional layers (Res-21, Res-22, Res-23, Res-24). We denote the deep feature extracted from feature map $I_t^{'}$ by this branch as $f_2\in\mathbb{R}^{D}$ ($D$ is dimension of feature vector $f_2$). Since convolutional neural layers are locally connected instead of fully connected, $f_2$ can only extract local feature and estimate the local value function of $I_t$, acting as a local guidance to planetary rovers.

The diagram of DB-CNN is illustrated in Fig.~\ref{Fig.3.1}, where Conv, Pool, Res, Fc and S are short for convolutional layer, max-pooling layer, residual convolution layer, fully-connected layer and softmax layer respectively. Compared with VIN, not only the depth of DB-CNN is reduced significantly, but also both global and local information of the image is kept and represented effectively. One typical parameter setting of DB-CNN is demonstrated in TABLE~\ref{tab3.1}.

\renewcommand\arraystretch{1.2}
\begin{table}[H]
\centering
\caption{Parameter setting of DB-CNN}\label{tab3.1}
\begin{tabular}{|c|c|c|}
\hline
\multirow{3}{*}{Reprocessing layers}&Conv-00&$6\times5\times5$ kernels with stride 1 \\ \cline{2-3}
~&Pool-00&$3\times3$ kernels with stride 2 \\ \cline{2-3}
~&Conv-01&$12\times4\times4$ kernels with stride 1 \\ \cline{2-3}
~&Pool-01&$3\times3$ kernels with stride 2 \\
\hline
\multirow{9}{*}{Branch one}&Conv-10&$20\times5\times5$ kernels with stride 1 \\ \cline{2-3}
~&Pool-10&$3\times3$ kernels with stride 2 \\ \cline{2-3}
~&Res-11&$20\times3\times3$ kernels with stride 1  \\ \cline{2-3}
~&Pool-11&$3\times3$ kernels with stride 2 \\ \cline{2-3}
~&Res-12&$20\times3\times3$ kernels with stride 1  \\ \cline{2-3}
~&Pool-12&$3\times3$ kernels with stride 1 \\ \cline{2-3}
~&Res-13&$20\times3\times3$ kernels with stride 1  \\ \cline{2-3}
~&Pool-13&$3\times3$ kernels with stride 1 \\ \cline{2-3}
~&Fc-1&192 nodes  \\ \cline{2-3}
~&Fc-2&10 nodes \\
\hline
\multirow{6}{*}{Branch two}&Conv-20&$20\times5\times5$ kernels with stride 1 \\ \cline{2-3}
~&Res-21&$20\times3\times3$ kernels with stride 1  \\ \cline{2-3}
~&Res-22&$20\times3\times3$ kernels with stride 1  \\ \cline{2-3}
~&Res-23&$20\times3\times3$ kernels with stride 1  \\ \cline{2-3}
~&Res-24&$20\times3\times3$ kernels with stride 1  \\ \cline{2-3}
~&Conv-21&$10\times3\times3$ kernels with stride 1 \\
\hline
\multirow{2}{*}{Output layers}&Fc-3&8 nodes \\ \cline{2-3}
~&S-1&8 nodes \\ \cline{2-3}
\hline
\end{tabular}
\end{table}

\subsection{Learning-based Global Path Planning Algorithm}
In this subsection, we illustrate the whole learning-based global path planning algorithm based on DB-CNN, which works as follows.
\subsubsection{Training Phase}
Since the expert dataset for global path planning is available, the training phase is offline. For each training step, we randomly choose one batch of data (line~3)  and calculate the loss $L(\alpha)$ according to Eq.~\label{eq2.9} (line~4). Then, we calculate the stochastic gradient $\nabla_{\alpha} L(\alpha)$ and update $\alpha$ through gradient descent with learning rate $\delta$ (line~5). A training epoch ends when all batches of data are employed to train for one time (line~2). After the number of training epoch reaches the maximum, the training phase will stop (line~1). 
\subsubsection{Planning Phase}
During the planning phase, satellite will firstly caputure the intial state $s_0$ including current orbital image $I_t$, the start position of planetary rover $(x^1_0,x^2_0)$, and the target position $(g^1_0,g^2_0)$ (line~1). Taking $s_0$ as input, DB-CNN will output the estimated value function $F_{\alpha}(s_0)$ (line~3). Hence, the moving direction for planetary rover $a_0$ can be determined according to $a_0=\pi(s_0)$ (line~4). After that, the position of planetary rover is changed into $(x^1_1,x^2_1)$ and the state can be updated into $s_1$ (line~5). By repeating this planning step until $(x^1_t,x^2_t)=(g^1_0,g^2_0)$ (line~2), the global path for panetary rover will be planned (as shown in the right part of Fig.~\ref{Fig.3.1}).
\subsubsection{Analysis of this Algorithm}
As shown in Fig.~\ref{Fig.3.1}, given the initial orbital image $I_0$ and target position of rover $(g^1_0,g^2_0)$, DB-CNN can output the estimated Q values of all positions through one forward calculation, since we can take the whole local feature map (output of layer Conv-21) as the partial input of layer Fc-3 directly. That is, the time and resource cost for calculating the Q value set $\{\hat Q(\{I_0, (g^1_0,g^2_0),(x^1,x^2)\},a)|(x^1,x^2)\in\mathcal{X}, a\in\mathcal{A}\}$ is approximately equal to the time and resource cost for calculating a single Q value $\hat Q(\{I_0, (g^1_0,g^2_0),(x^1_0,x^2_0)\},a)$. Therefore, the planning loop (line 3-5) during online planning phase only requires computation at the initial step. Most significantly, when multiple rovers distributed in different places share the same destination, traditional search algorithms (e.g. A*) have to plan path for each rover one by one. By contrast, DB-CNN is capable of planning paths for them simultaneously through one forward calculation, the efficiency of which is enhanced significantly. 
\begin{algorithm}[!t]
\caption{Learning-based global path planning algorithm}
\label{algor-3-2}
\textbf{Training phase (offline)}
\begin{algorithmic}[1]
\For{epoch in $1,\cdots,K$} 
\For{step in $1,\cdots, M$}
\State Randomly select one batch of training data.
\State Calculate the loss $L(\alpha)$ and its gradient $\nabla_{\alpha} L(\alpha)$ based on Eq.~(\ref{eq2.7}).
\State Update DB-CNN $F_{\alpha}$ by $\alpha^{'} \leftarrow \alpha-\delta\nabla_{\alpha} L(\alpha)$.
\EndFor
\EndFor
\end{algorithmic}
\textbf{Planning phase (online)}
\begin{algorithmic}[1]
\State Receive the initial state $s_0=\{I_0,(g^1_0,g^2_0),(x^1_0,x^2_0)\}$.
\While{$(x^1_t,x^2_t)\ne(g^1,g^2)$}
\State Input $s_t$ into DB-CNN and output $F_{\alpha}(s_t)$.
\State Choose action $a_t$ based on $F_{\alpha}(s_t)$ and Eq.~(\ref{eq2.8}).
\State Update the state $s_t$ into $s_{t+1}$ (Note: $\forall t\geq 0, I_t=I_0$, $(g^1_{t},g^2_{t})=(g^1_{0},g^2_{0})$).
\EndWhile
\State Send global path planning results to planetary rover.
\end{algorithmic}
\end{algorithm}
\section{Experiments and Analysis}
\subsection{Experimental Settings}
We evaluate the planetary global path planning performace of the proposed DB-CNN on two datasets as follows.

1) \textbf{Grid maps with obstacles}. It is composed of 10000 grid maps with size 64 $\times$ 64 and random obstacles, where 0 represents free grid and 1 represents obstacle. Each input consists of one grid map, one target map, and positions of the rover. Since grid maps can be viewed as simplified planetary orbital images, this dataset has been widely used to evaluate global path planning algorithms.

\begin{figure}[H]
\centering
\subfigure[Grid maps]{
\includegraphics[width=0.7\linewidth]{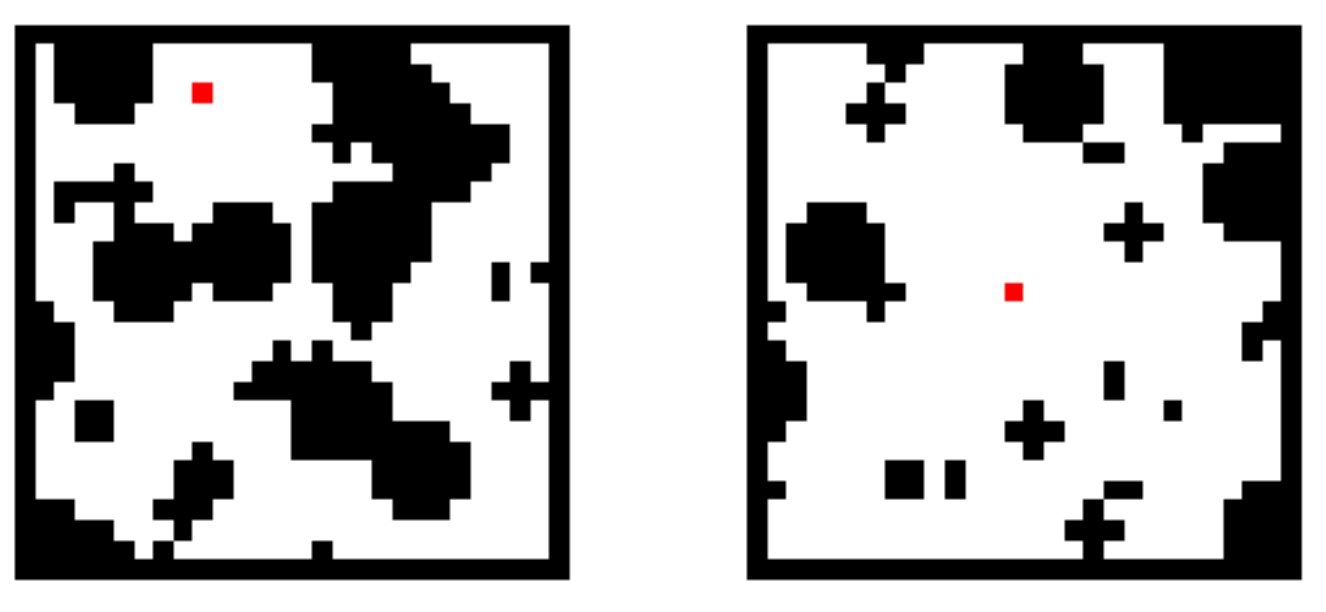}}
\subfigure[Grey image]{\label{Fig.4.0}
\includegraphics[width=0.31\linewidth]{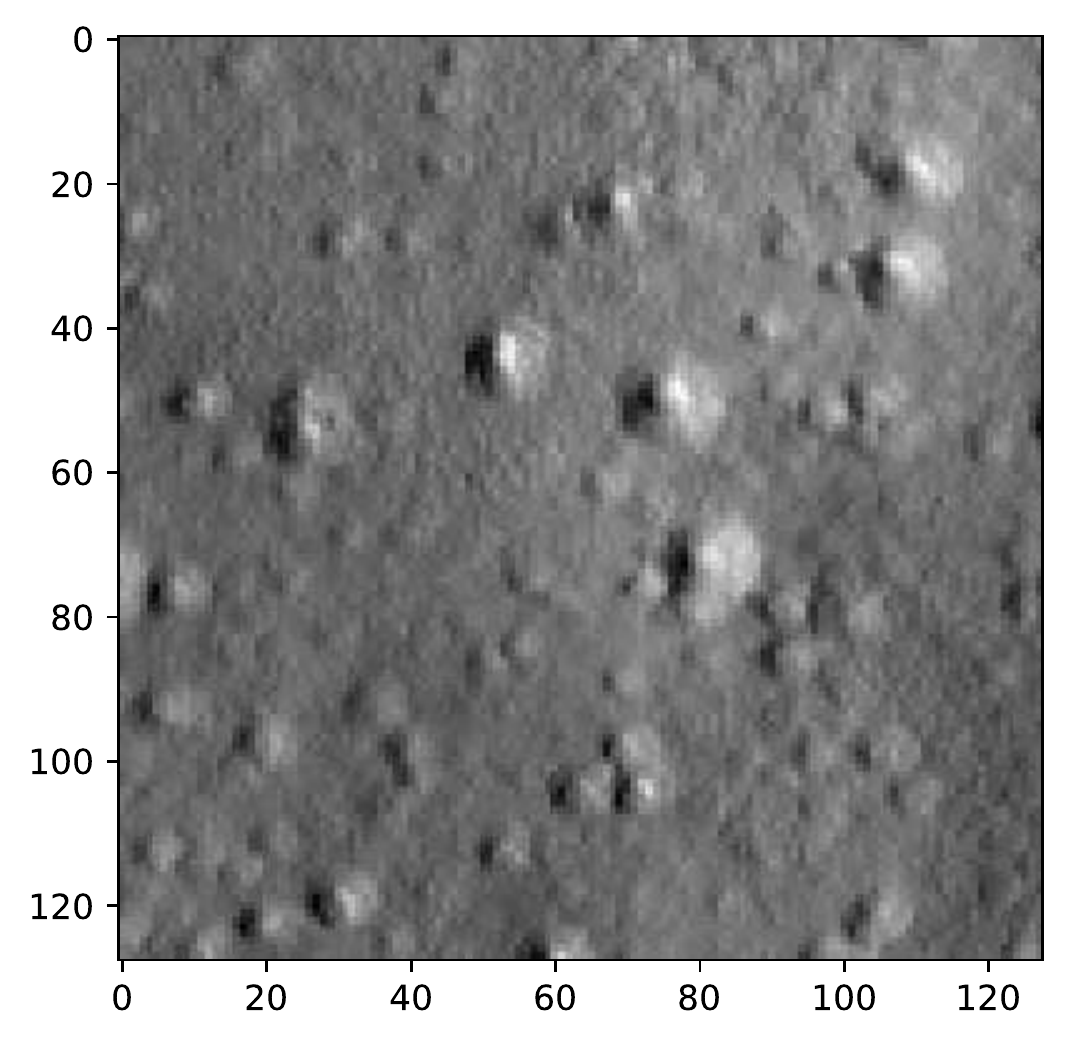}}
\subfigure[Edge image]{
\includegraphics[width=0.31\linewidth]{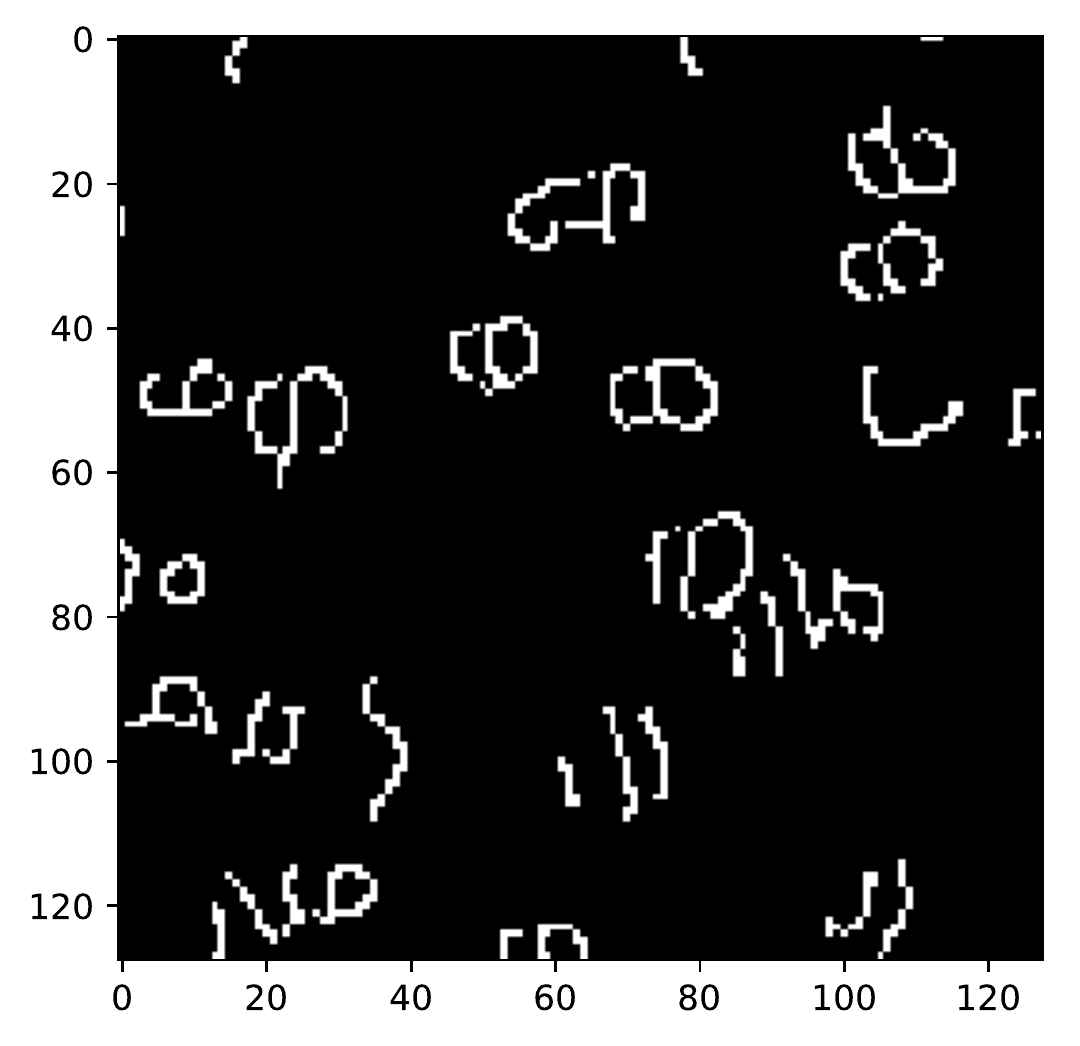}}
\subfigure[Target image]{
\includegraphics[width=0.31\linewidth]{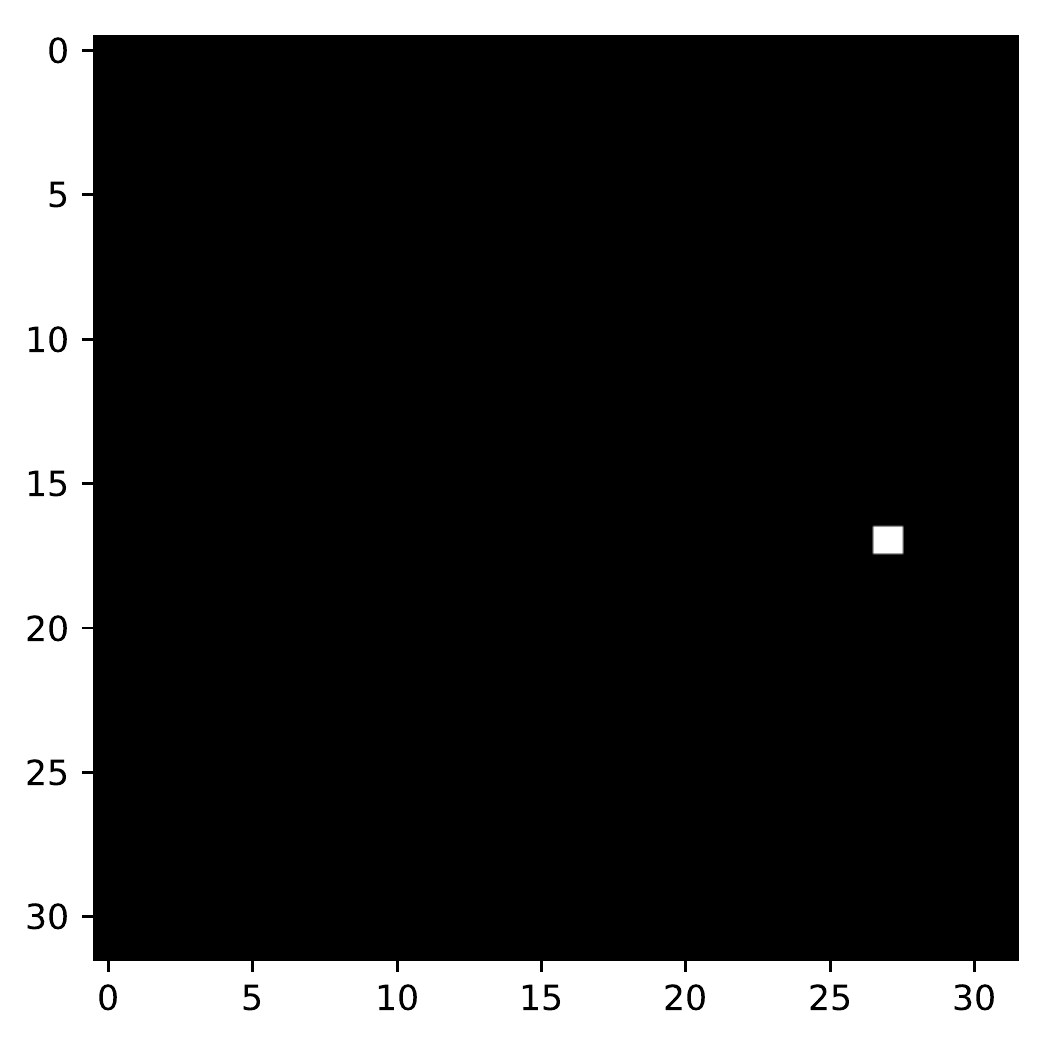}}
\caption{Examples for two dataset.}
\end{figure}

2) \textbf{Martian surface images from HiRISE} \cite{ref24}. This dataset is generated from high-resolution Martian images captured by real orbit detectors, which consists of 10000 images with size 128 $\times$ 128. Each input consists of one gray image of Martian surface, one edge image generated by Canny algorithm \cite{ref25} for edge augmentation, one target image, and input positions of planetary rovers. We choose Martian surface images because they exhibit typical features among explorable planets such as craters with various sizes (as shown in Fig.~\ref{Fig.4.0}). The global path planning algorithm based on DB-CNN can also be extended into other planetary scenarios. 

For each dataset, the outputs of each input are the optimal moving directions of all given positions, and we randomly choose 6/7 data for training and the remaining 1/7 data for testing. 
\subsection{Compared Baseline Architectures}
We compare DB-CNN with three CNN baselines as follows.

1) \textbf{VIN}. This is the state-of-the-art deep neural network structure on path planning with fully observations. The parameter settings are the same as those in \cite{ref14} and the iteration number K in VIN is set as 80.

2) \textbf{ResNet}. This is a classical residual network, which keeps branch two of DB-CNN while deletes branch one of DB-CNN. By comparing DB-CNN with ResNet, we can evaluate whether branch one of DB-CNN enhances the global path planning accuracy for rovers.

3) \textbf{DCNN}. This is a common CNN comprised of convolutional layers, max-pooling layers and fully connected layers, which is also modified from branch two of DB-CNN. However, compared with ResNet, it replaces residual layers with basic convolutional layers.

The metrics we employed to evaluate their performace on global path planning tasks are global path planning accuracy and global path planning success rate, where accuracy is defined as the percentage of optimal moving direction predicted by them and success rate is defined as the percentage of safe paths planned by them. 

\subsection{Results and Discussions}
The training performance of all architectures on two datasets are shown in Fig.~\ref{Fig.4.1} and the final experimental results are reported in TABEL~\ref{tab.4.1}.
\subsubsection{Training Results Analysis}
\begin{figure}
\centering
\includegraphics[width=0.49\linewidth]{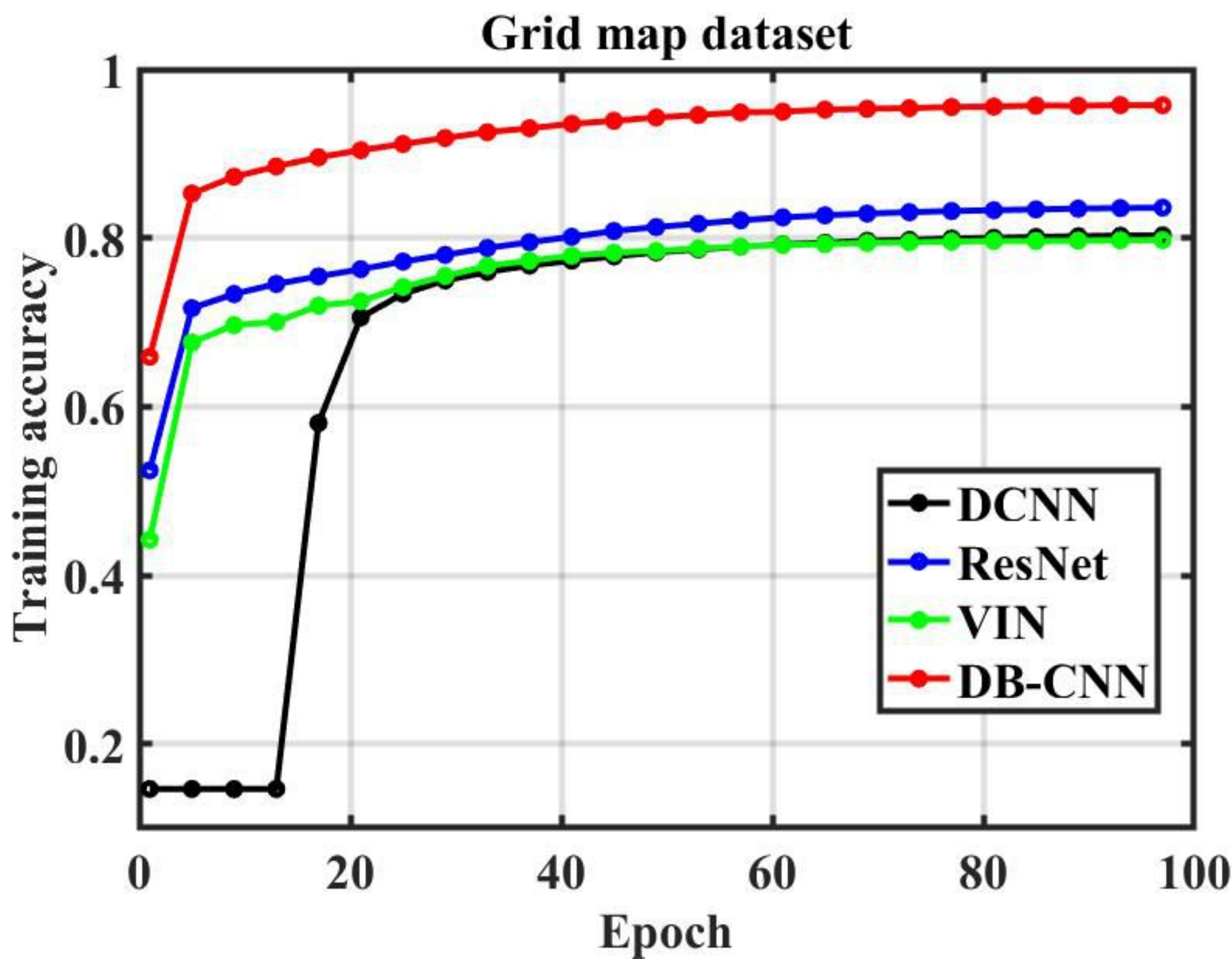}
\includegraphics[width=0.49\linewidth]{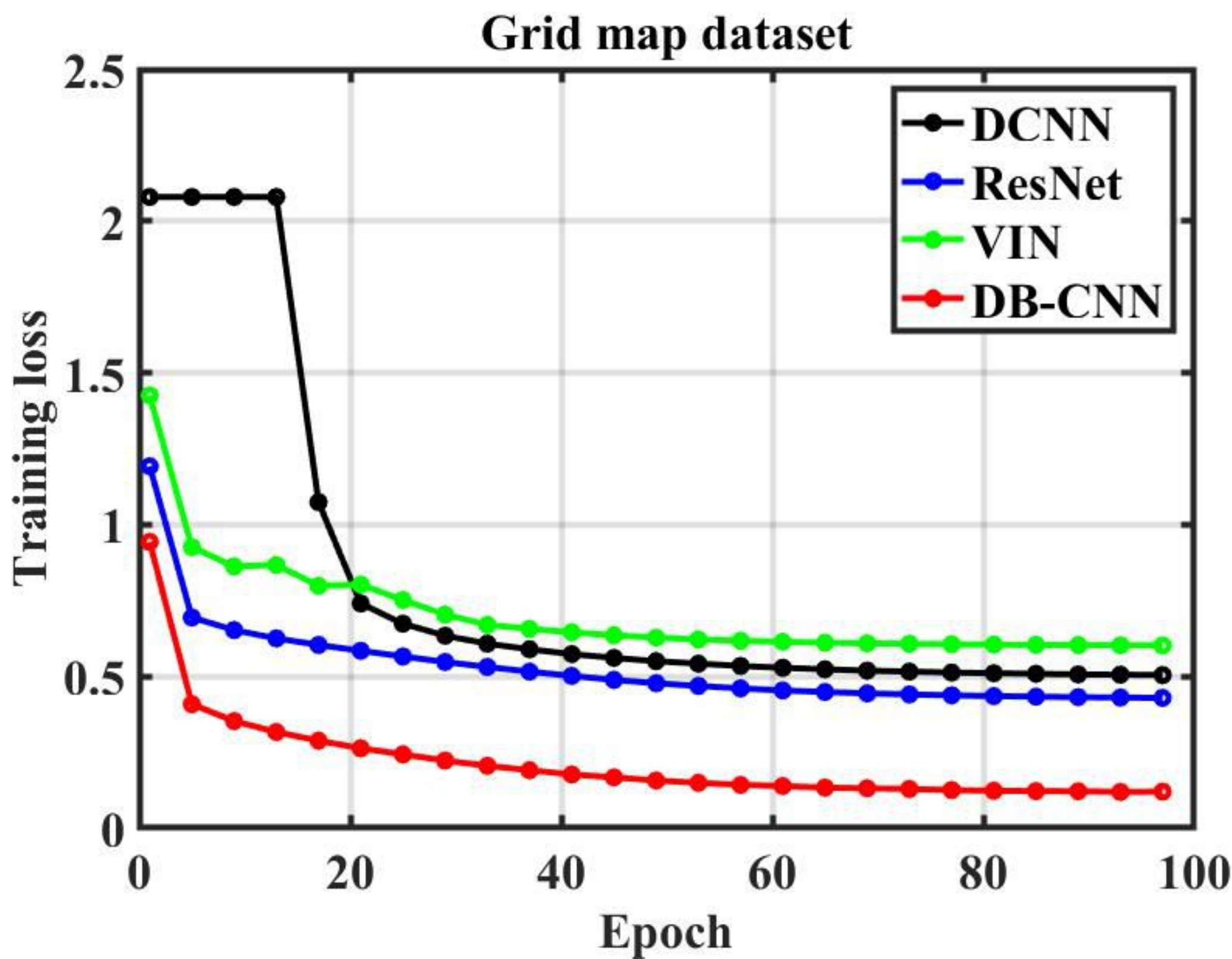}
\includegraphics[width=0.49\linewidth]{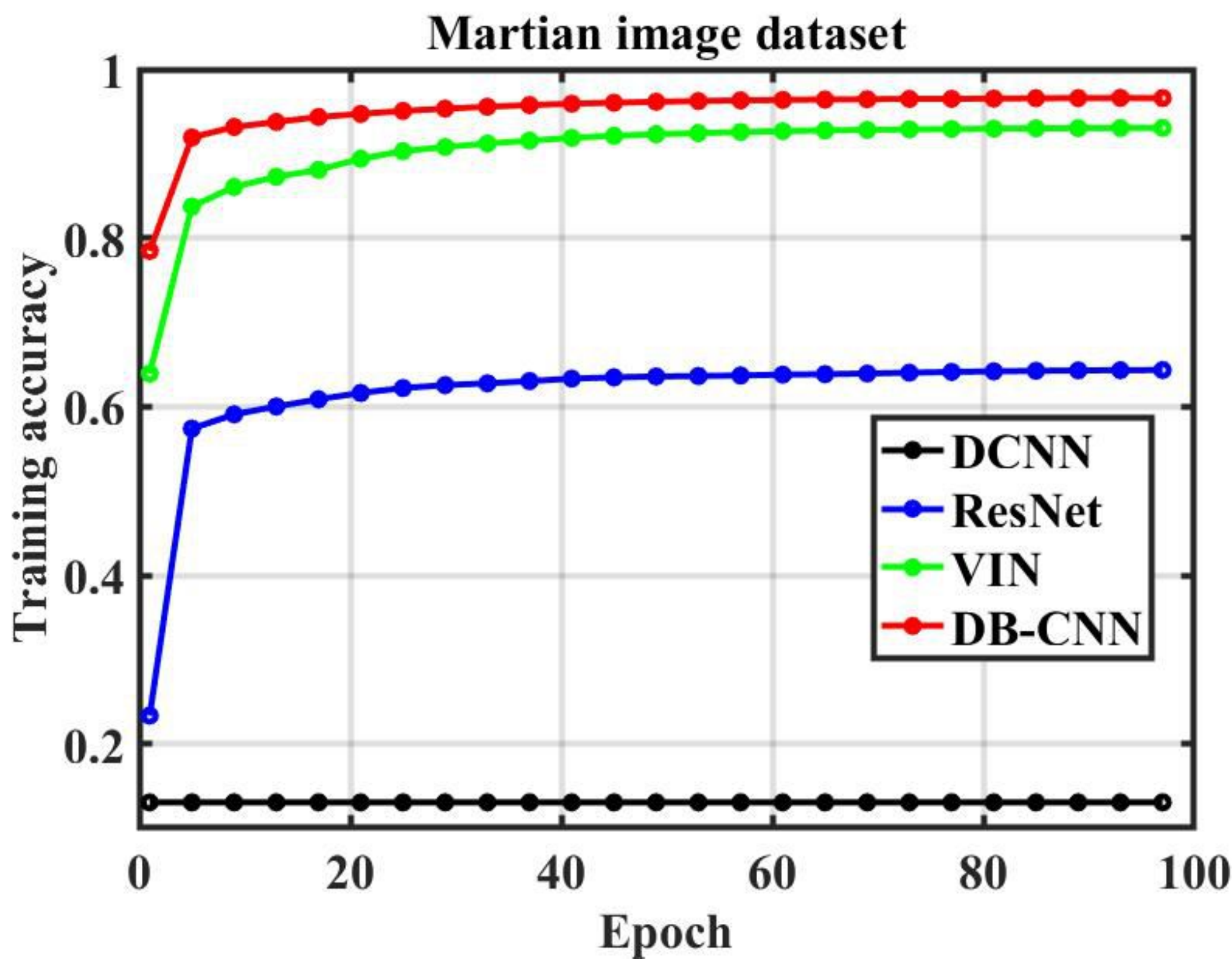}
\includegraphics[width=0.49\linewidth]{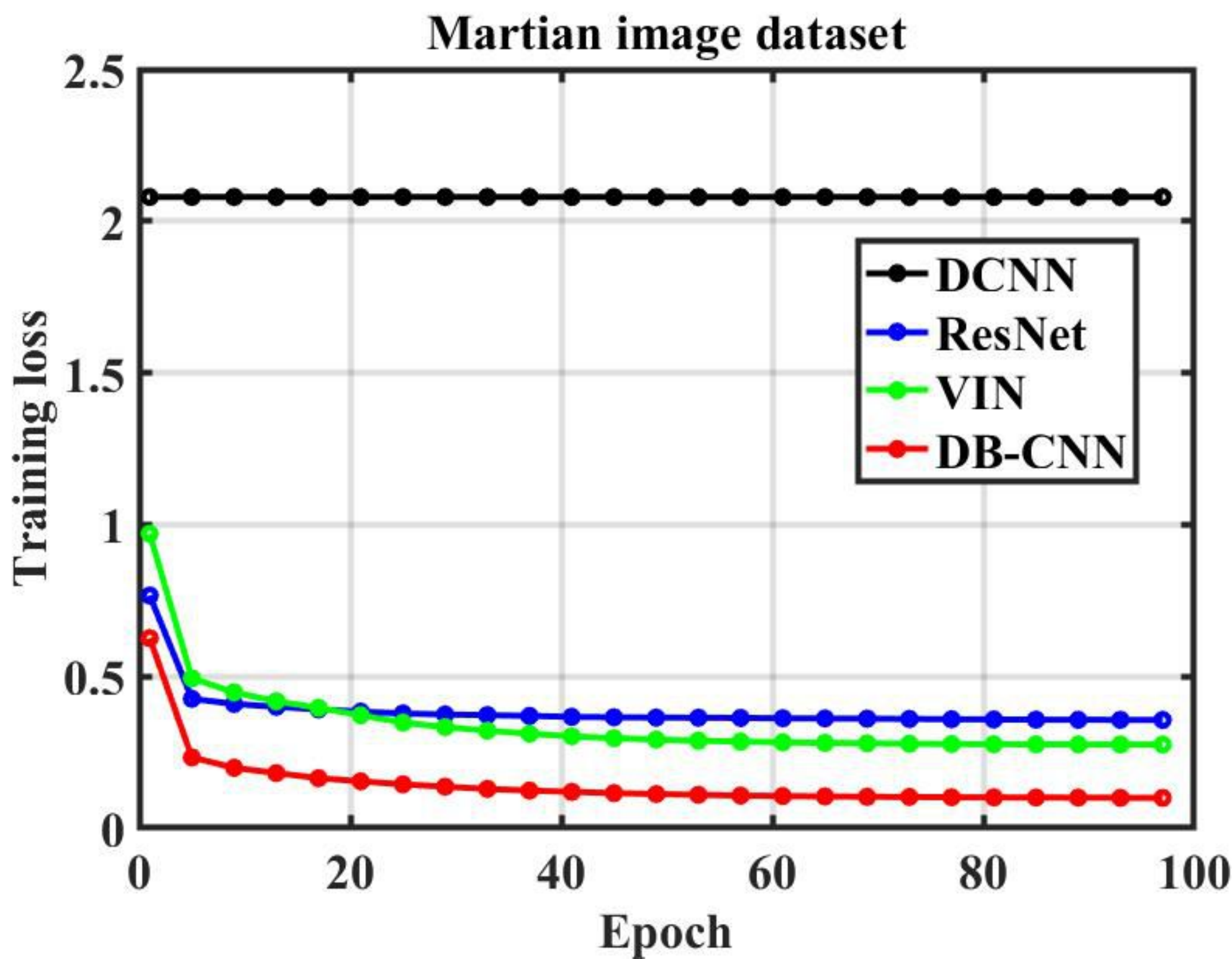}
\caption{Training performance of all architectures on two datasets.}\label{Fig.4.1}
\end{figure}

As illustrated in Fig.~\ref{Fig.4.1}, both training accuracy and training loss of DB-CNN converge faster than other baseline CNNs. After 100 training epoches, DB-CNN achieves both the higher Acc1 and the higher SR1 on all datasets, outperforming other baselines CNNs significantly (as shown in TABLE~\ref{tab.4.1}). Moreover, compared with the state-of-the-art architecture---VIN, the training time of DB-CNN is largely reduced, which means that DB-CNN has more efficient structure that VIN. Notably, the training time of ResNet and DCNN is smaller than DB-CNN because they only keep partial structure of DB-CNN. Therefore, it can be conclude that DB-CNN is a more accurate and efficient architecture for planning path directly from planetary orbital images.
\subsubsection{Testing Results Analysis}
As reported in TABLE~\ref{tab.4.1}, DB-CNN also keeps its superior global path planning performance on testing data. Remarkably, the planetary orbital images in testing data are totally different from those in training data, which demonstrates that DB-CNN is capable of planning path from unknown planetary orbital images after training. Since planetary rovers are commonly required to explore unknown environments, the algorithms also need to plan path from unknown planetary environments. Hence, DB-CNN is more effective for planning path from planetary orbital images in practice compared with other baselines.

\begin{figure*}
\centering
\begin{minipage}[b]{0.24\linewidth}
\centering
\includegraphics[width=1\linewidth]{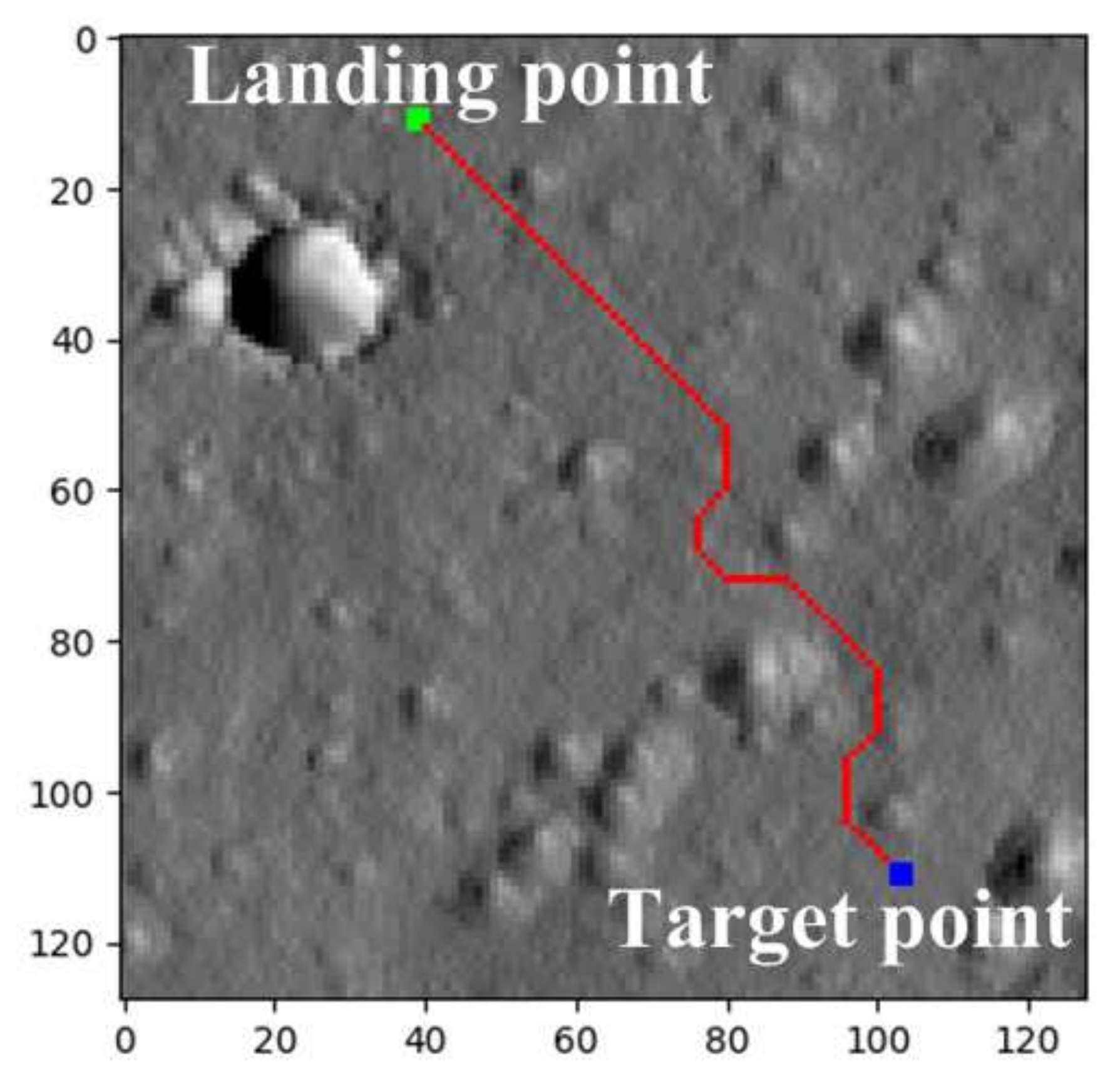}\\
\includegraphics[width=1\linewidth]{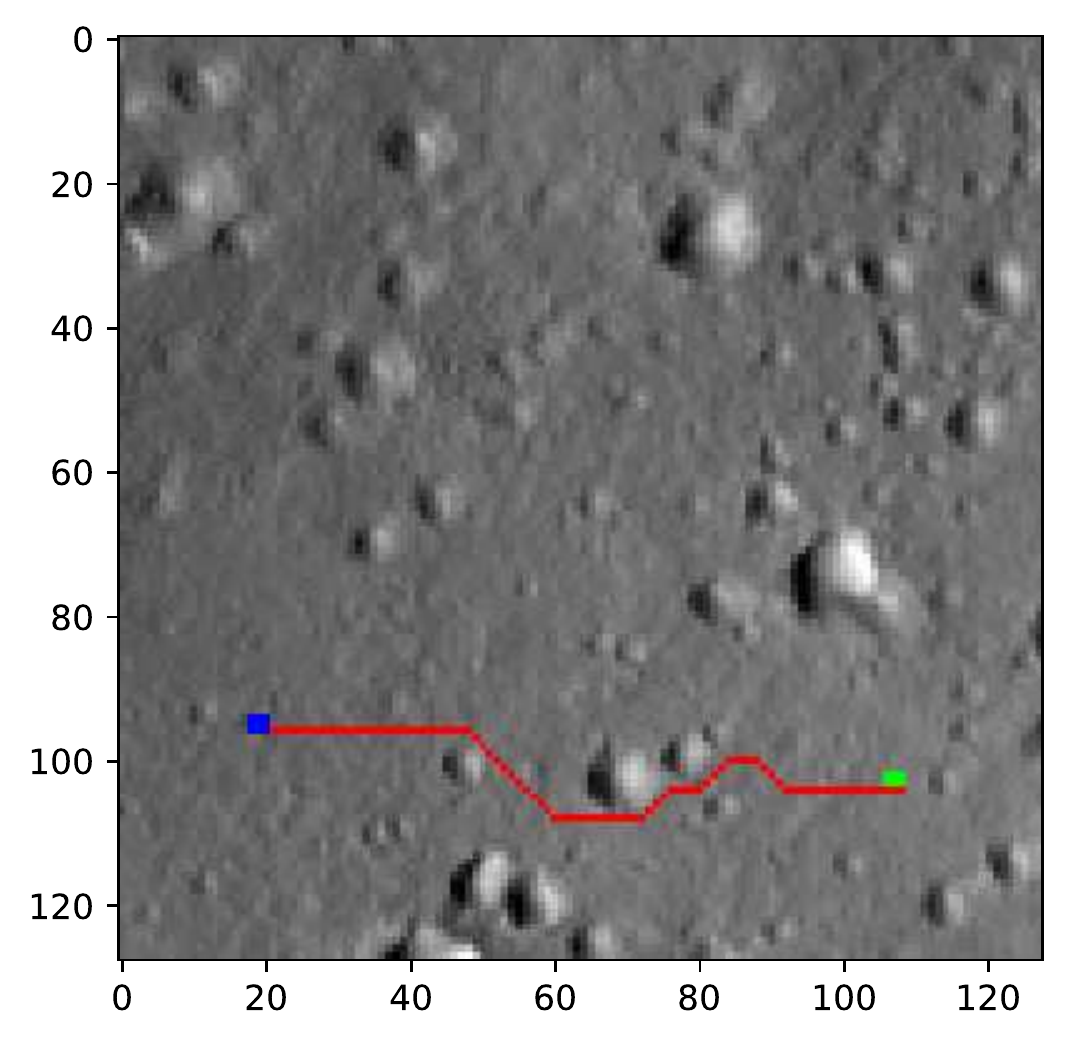}\\
\end{minipage}
\begin{minipage}[b]{0.24\linewidth}
\centering
\includegraphics[width=1\linewidth]{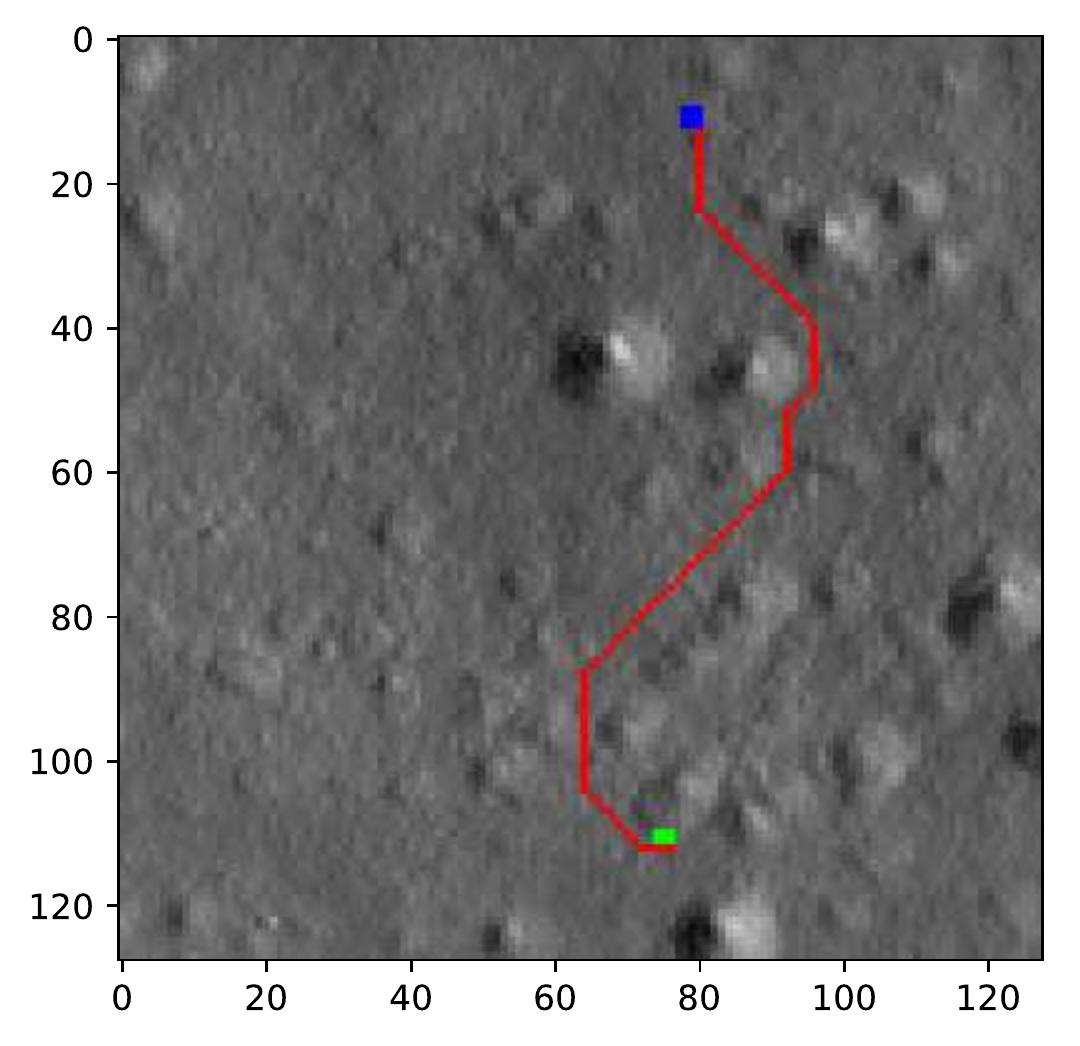}\\
\includegraphics[width=1\linewidth]{figure/Martians/success66}\\
\end{minipage}
\begin{minipage}[b]{0.24\linewidth}
\centering
\includegraphics[width=1\linewidth]{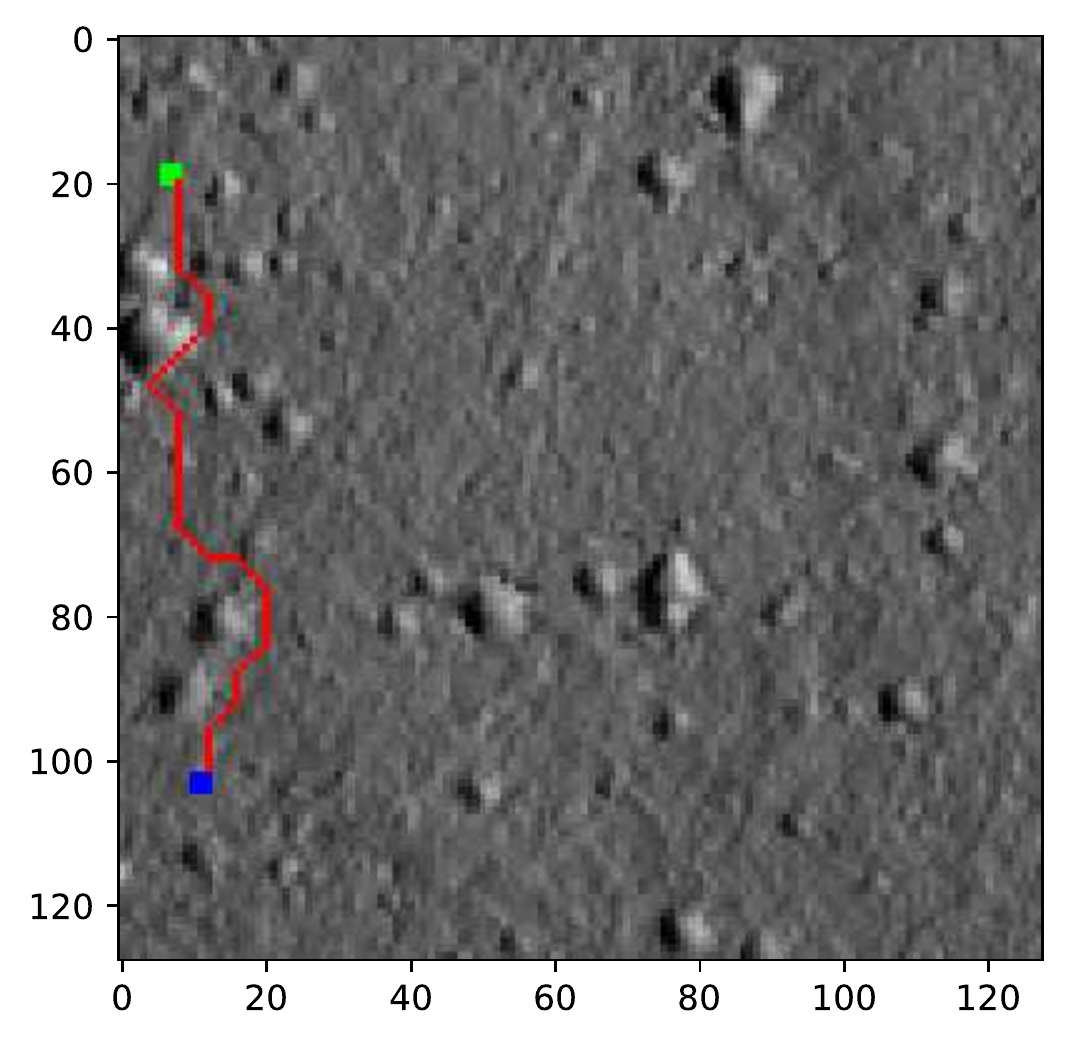}\\
\includegraphics[width=1\linewidth]{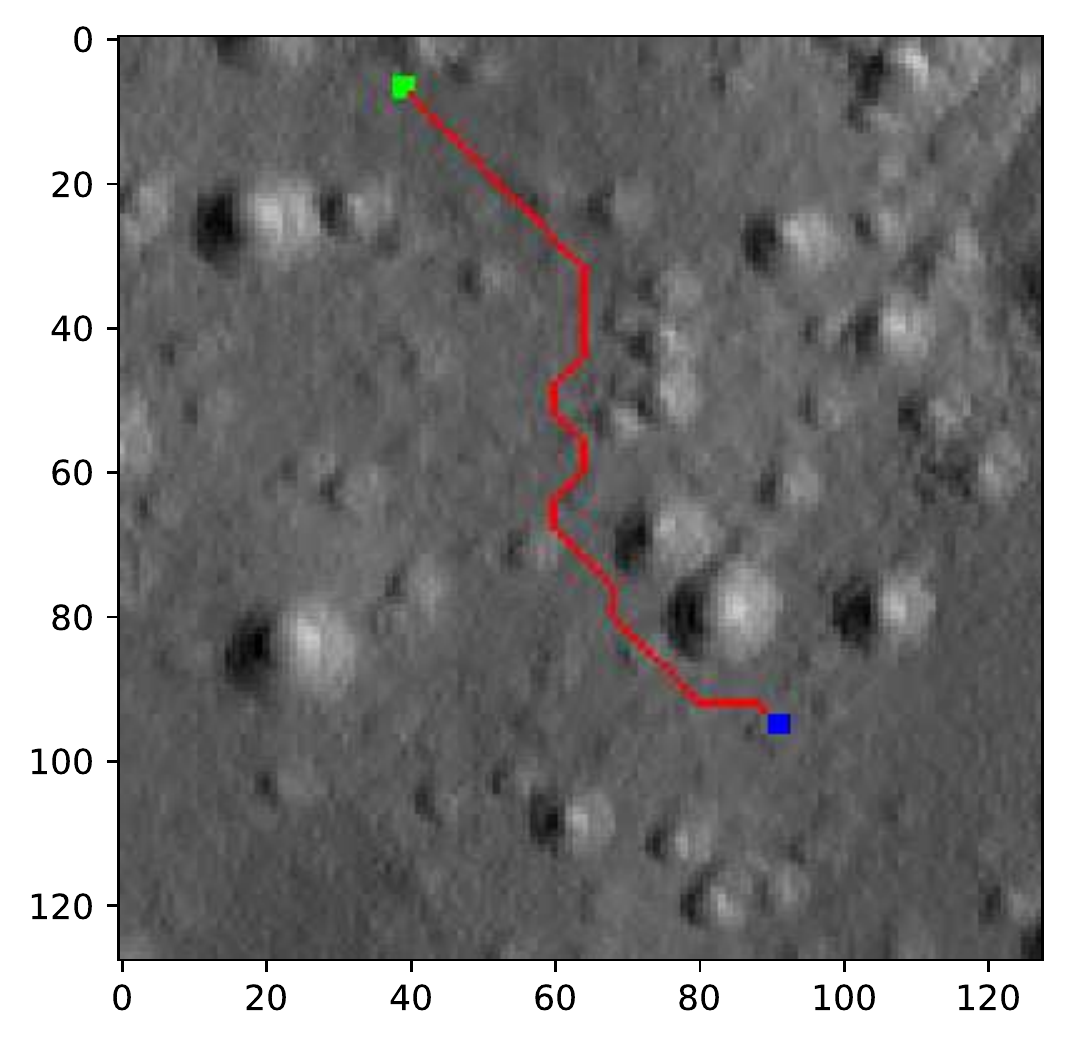}
\end{minipage}
\begin{minipage}[b]{0.24\linewidth}
\centering
\includegraphics[width=1\linewidth]{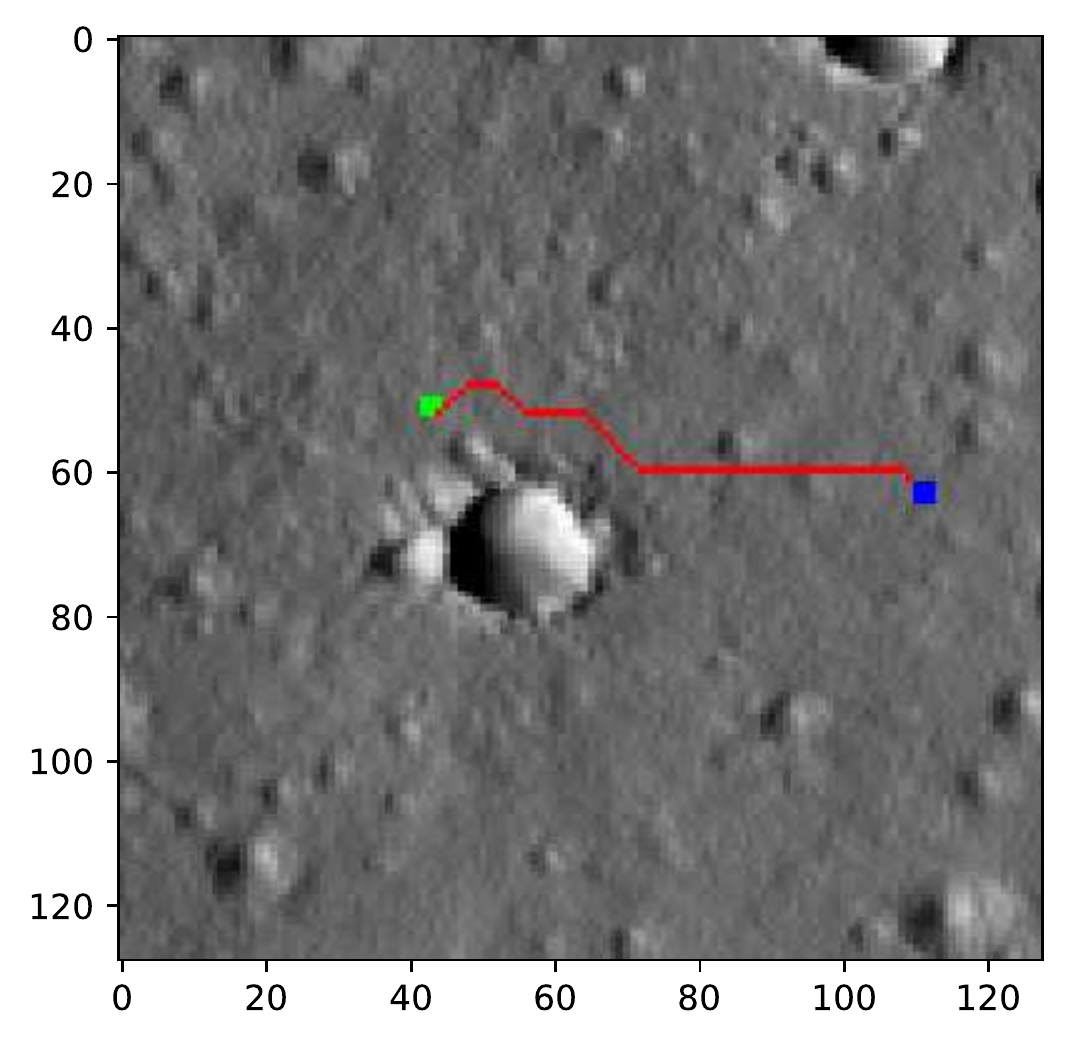}\\
\includegraphics[width=1\linewidth]{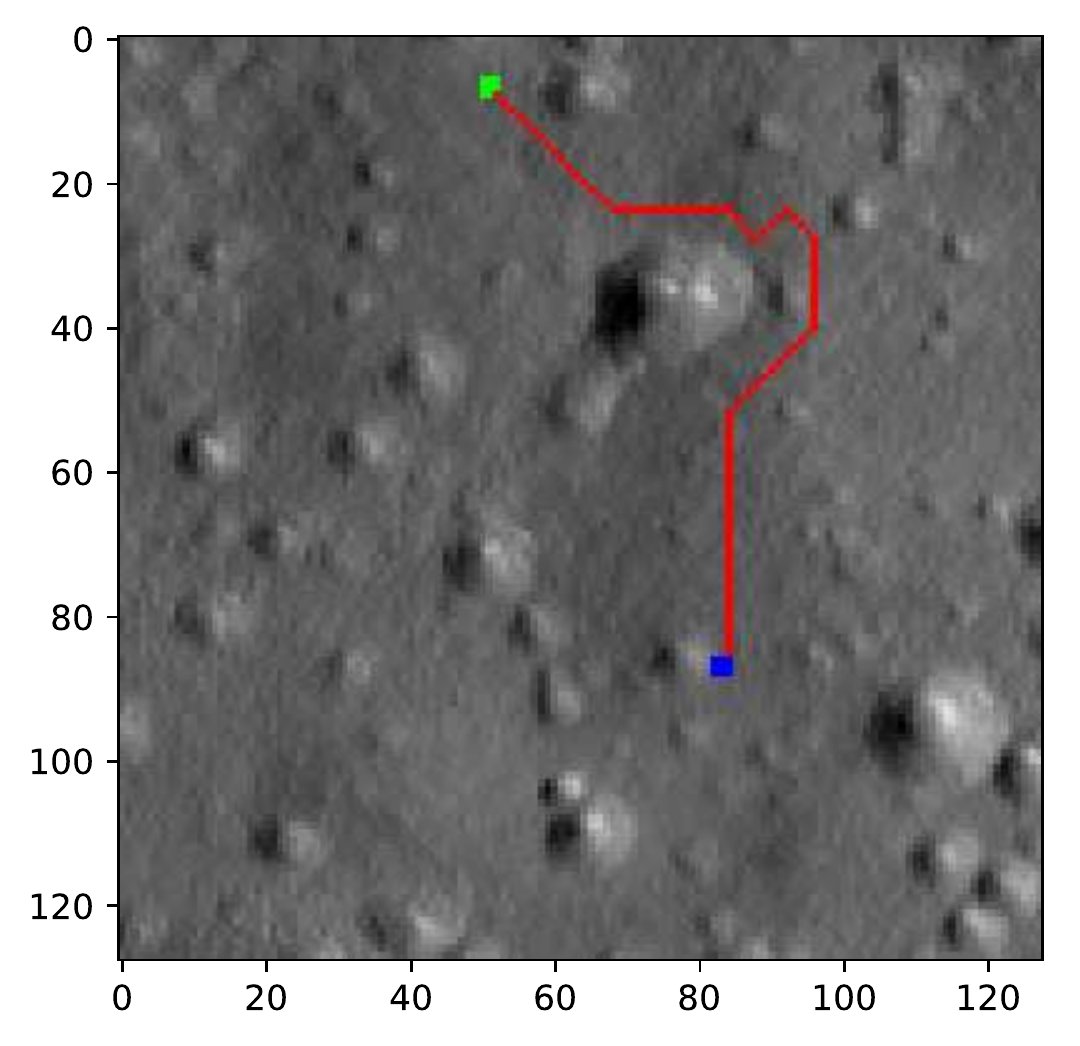}
\end{minipage}
\caption{Experiments on $128\times128$ Martian images. (Green points are the landing points. Blue points are the target points. Navigation trajectories are red.)}\label{Fig.5.1}

\end{figure*}

Fig.~\ref{Fig.5.1} presents some successful path planned by DB-CNN from Martian orbital images. It can be seen that the paths for rover avoid craters with varying size precisely under the guidance of DB-CNN. Furthermore, the trajectories are nearly optimal. It is noteworthy that prior knowledge of craters are unknown and DB-CNN has to learn and understand these deep features of original Martian images through training. Therefore, the performance of DB-CNN is marvellous.

\begin{table}
\centering
\caption{Accuracy and success rate of all architectures on two datasets.}\label{tab.4.1}
\begin{threeparttable}
\begin{tabular}{|c|c|c|c|c|c|}
\hline 
Dataset & Metrics & DB-CNN & VIN & ResNet & DCNN \\
\hline
\multirow{5}{*}{Grid maps}     & Acc1 & \textbf{93.8\%} & 80.3\% & 83.6\% & 80.3\%\\\cline{2-6}
                             ~ & Acc2 & \textbf{88.5\%} & 80.8\% & 73.1\% & 76.4\%\\\cline{2-6}           
                             ~ & SR1 & \textbf{94.7\%} & 47.5\% & 40.0\% & 39.9\%\\
\cline{2-6}
                             ~ & SR2 & \textbf{80.2\%} & 49.5\% & 33.8\% & 37.5\%\\
\cline{2-6}
                             ~ &ET & 25.0s & 56.4s   & 19.9s   & 19.2s \\                             
\hline

\multirow{5}{*}{Martian images}& Acc1 & \textbf{96.5\%} & 93.1\% & 87.4\% & 13.0\%\\\cline{2-6}
                             ~ & Acc2 & \textbf{96.5\%} & 93.0\% & 86.1\% & 12.7\%\\\cline{2-6}           
                             ~ & SR1  & \textbf{96.3\%} & 83.7\% & 69.0\% & 1.1\%\\
\cline{2-6}
                             ~ & SR2  & \textbf{92.3\%} & 83.8\% & 67.5\% & 1.3\%\\
\cline{2-6}
                             ~ & ET   & 53.4s  &151.6s  &41.0s  & 40.8s \\                             
\hline
\end{tabular}
\begin{tablenotes}
\footnotesize
\item[1] Acc1: global path planning accuracy on training data. 
\item[2] Acc2: global path planning accuracy on testing data. 
\item[3] SR1: global path planning successful rate on training data. 
\item[4] SR2: global path planning successful rate on testing data. 
\item[5] ET: the time cost for each training epoch.
\end{tablenotes}
\end{threeparttable}
\end{table}
\begin{figure}
\subfigure[Original Martian images]{
\begin{minipage}[c]{0.24\textwidth}
\centering
\includegraphics[width=1\linewidth]{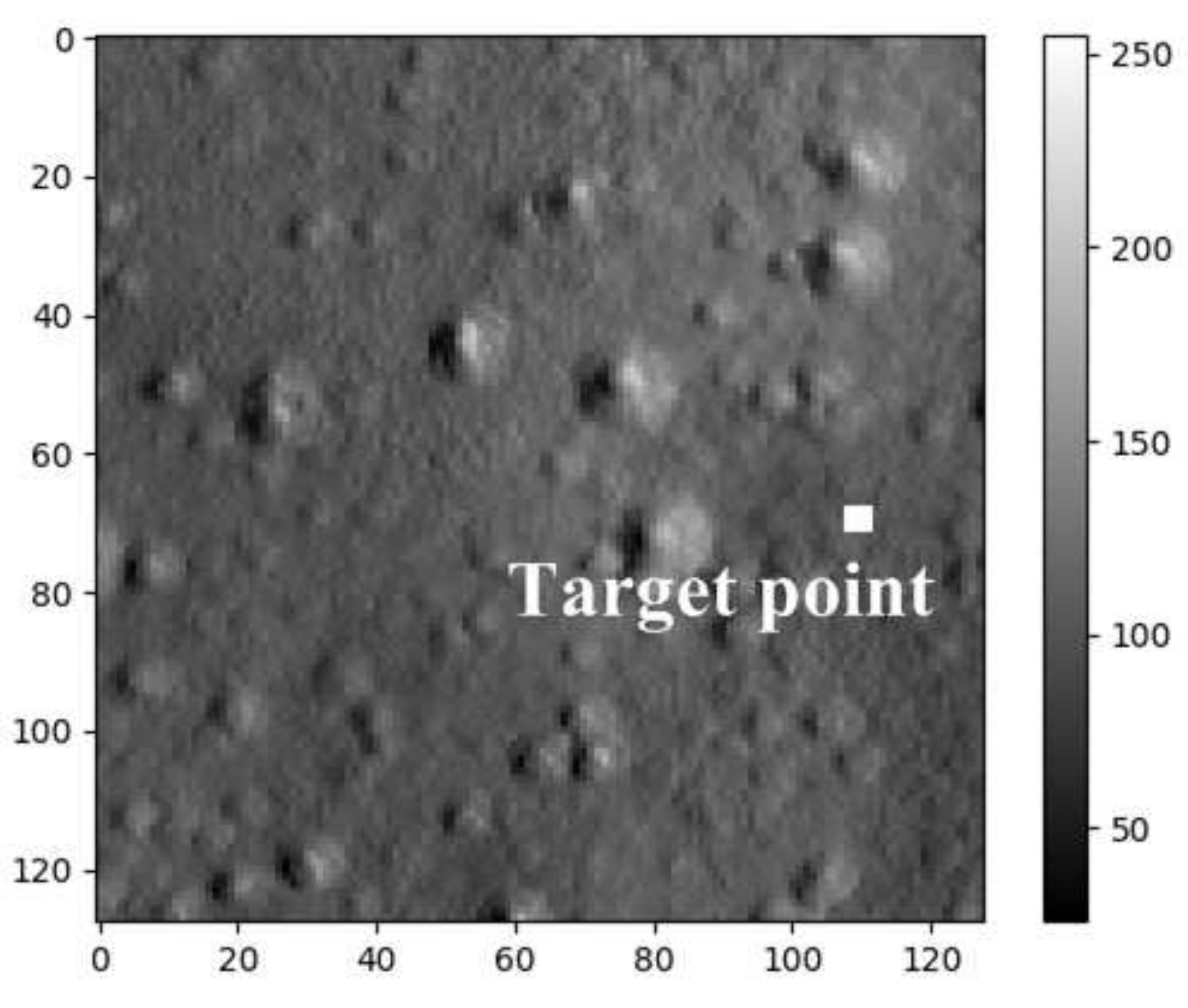}
\end{minipage}
\begin{minipage}[c]{0.24\textwidth}
\centering
\includegraphics[width=1\linewidth]{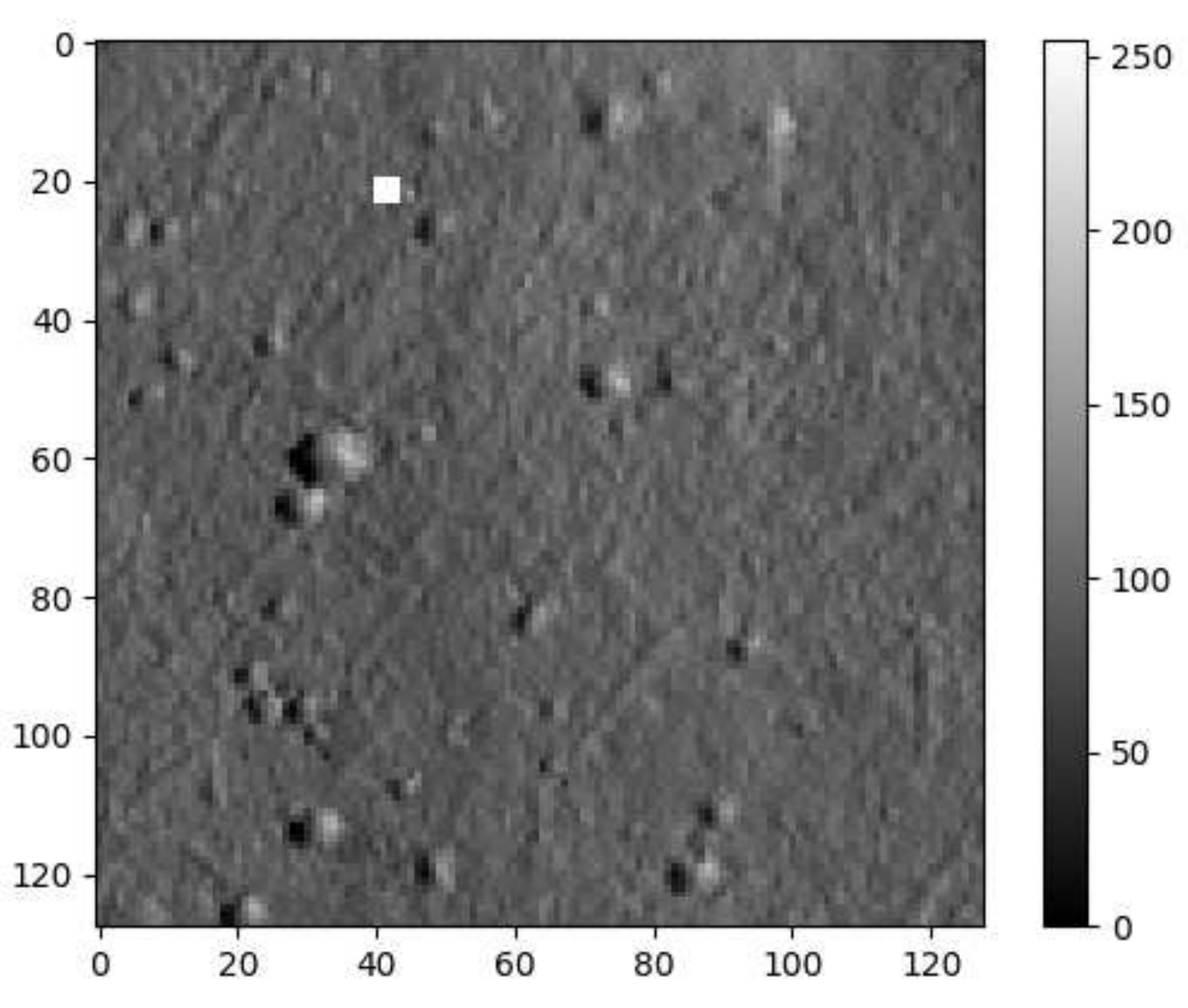}
\end{minipage}}
\subfigure[Value functions estimated by DB-CNN]{
\begin{minipage}[c]{0.24\textwidth}
\centering
\includegraphics[width=1\linewidth]{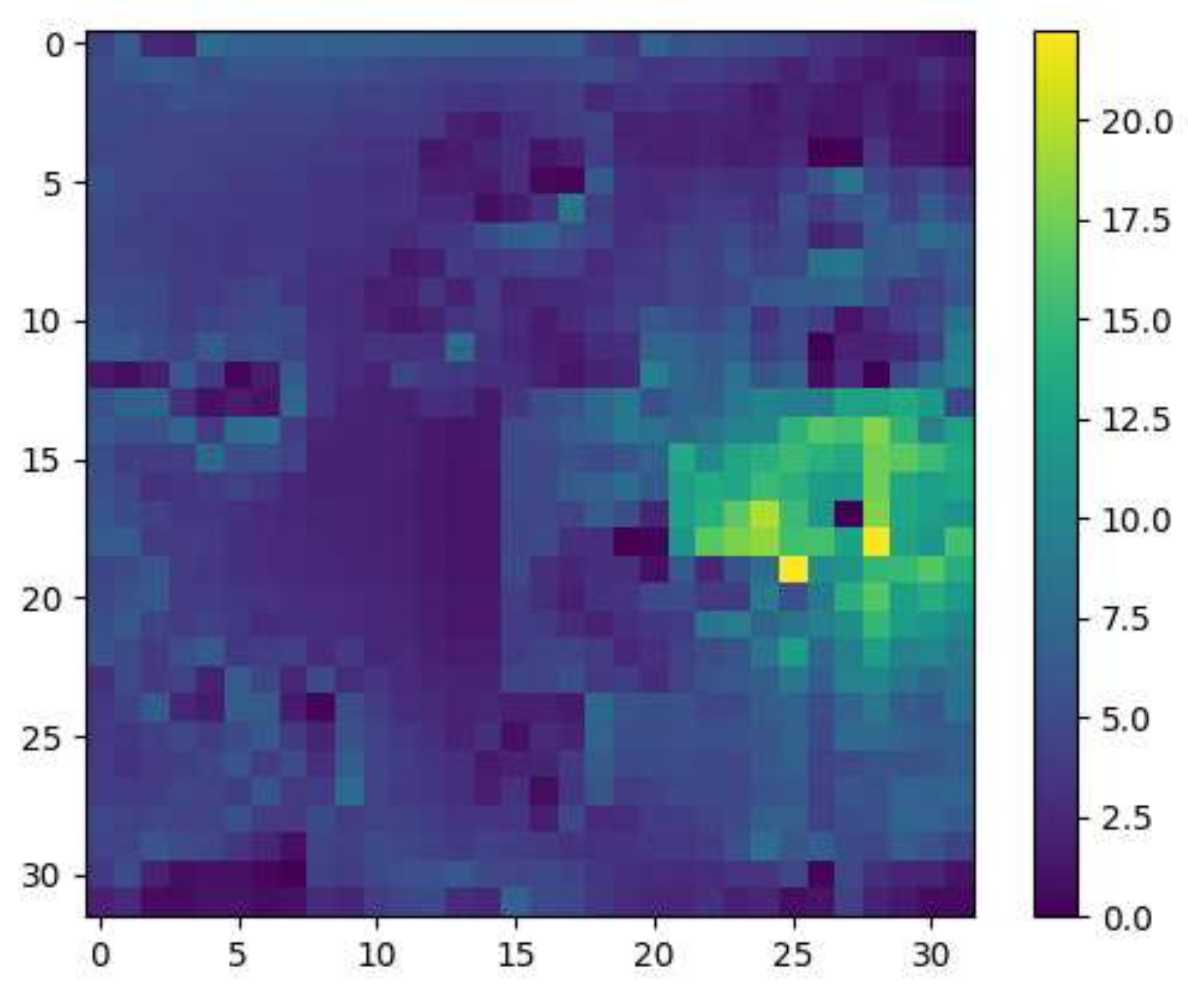}
\end{minipage}
\begin{minipage}[c]{0.24\textwidth}
\centering
\includegraphics[width=1\linewidth]{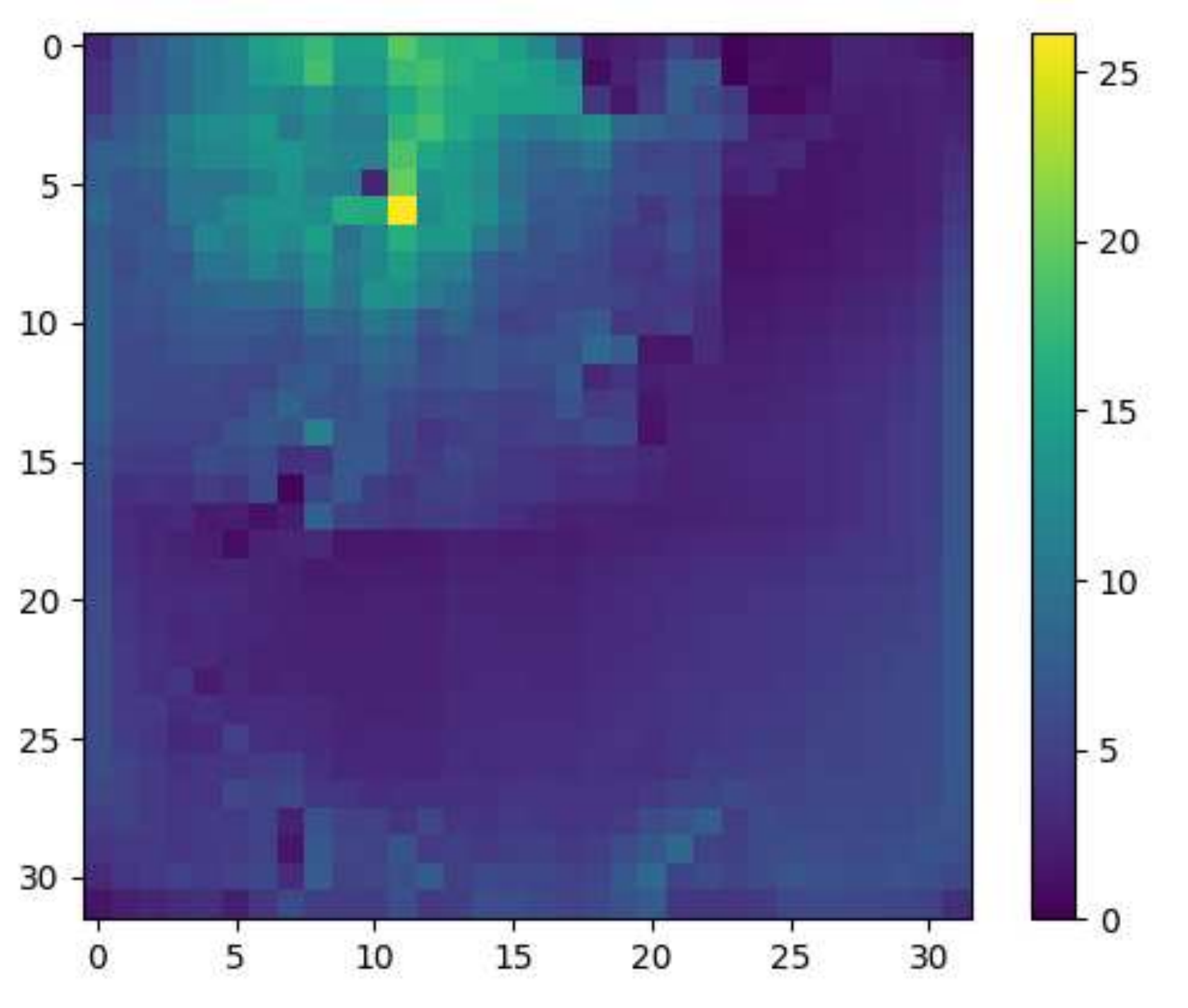}
\end{minipage}}
\subfigure[Value functions estimated by ResNet]{
\begin{minipage}[c]{0.24\textwidth}
\centering
\includegraphics[width=1\linewidth]{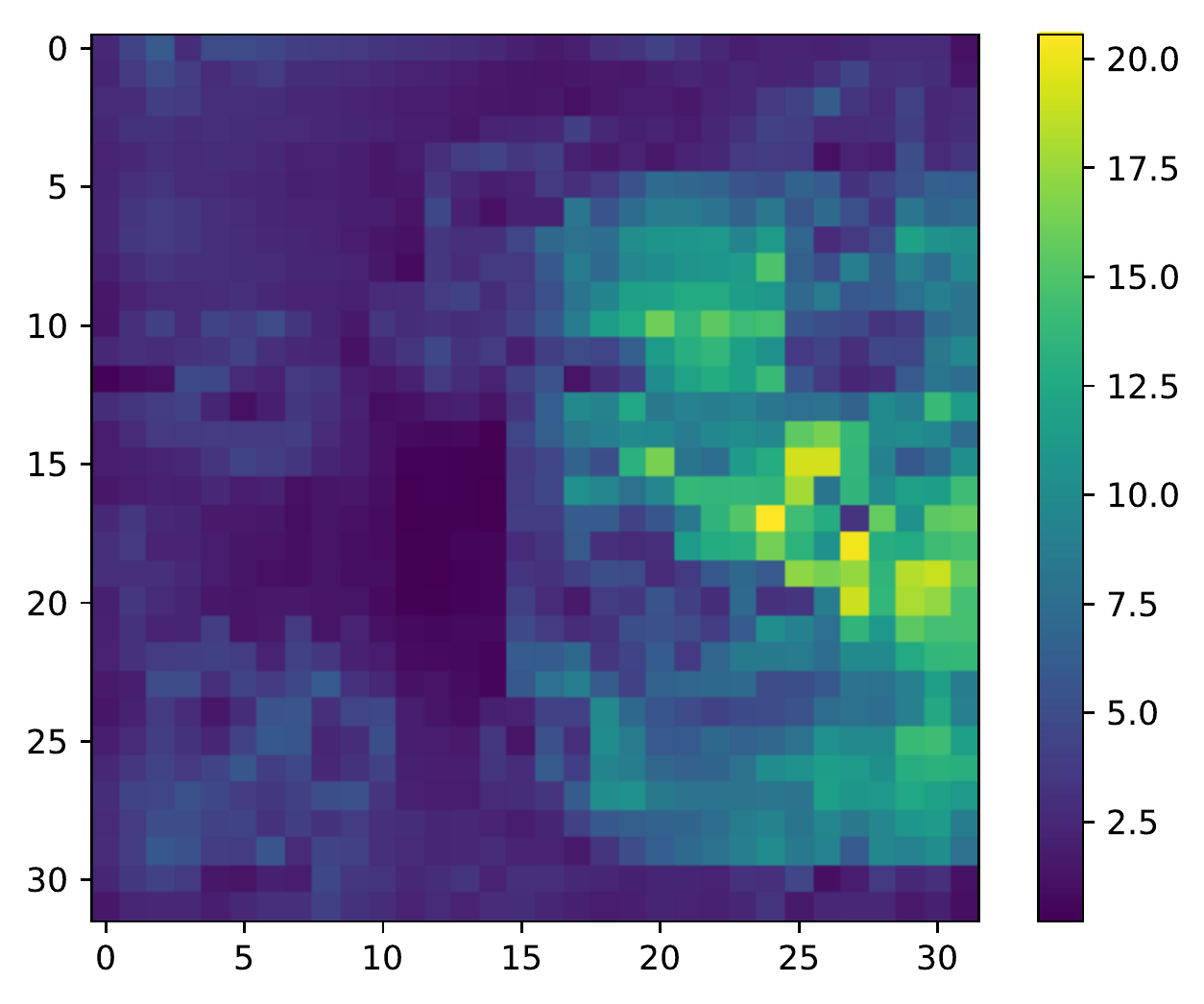}
\end{minipage}
\begin{minipage}[c]{0.24\textwidth}
\centering
\includegraphics[width=1\linewidth]{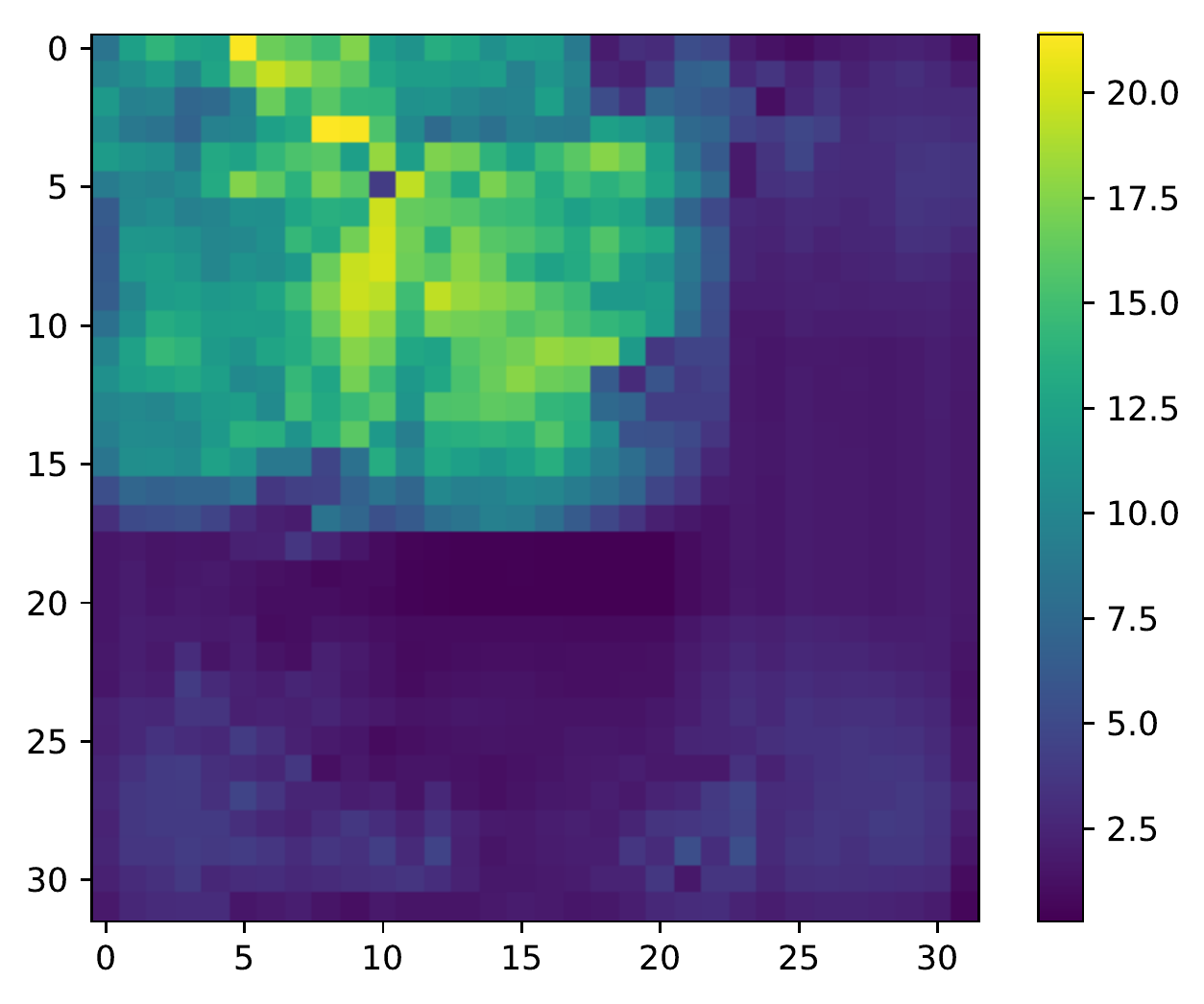}
\end{minipage}}
\subfigure[Value functions estimated by VIN]{
\begin{minipage}[c]{0.24\textwidth}
\centering
\includegraphics[width=1\linewidth]{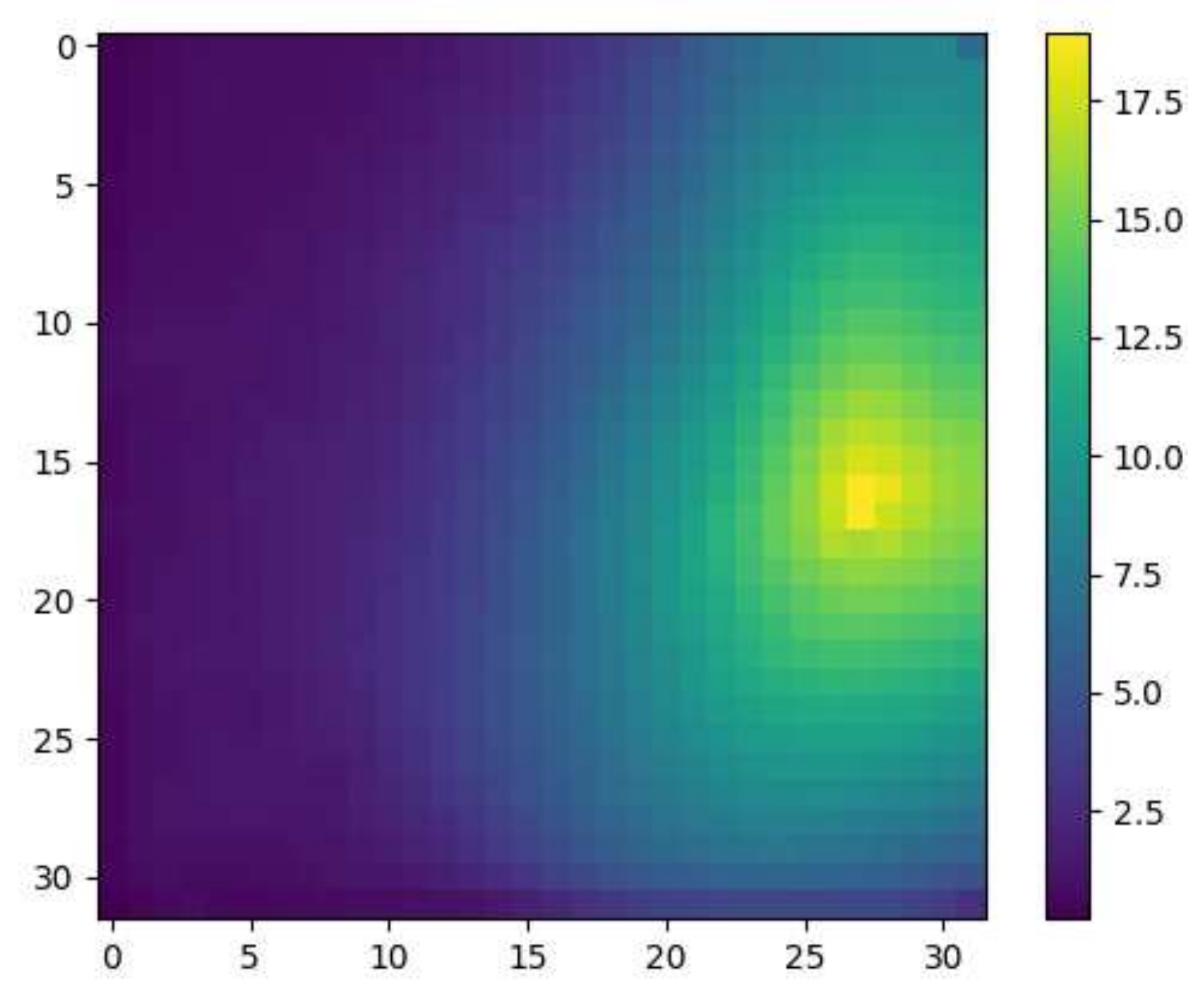}
\end{minipage}
\begin{minipage}[c]{0.24\textwidth}
\centering
\includegraphics[width=1\linewidth]{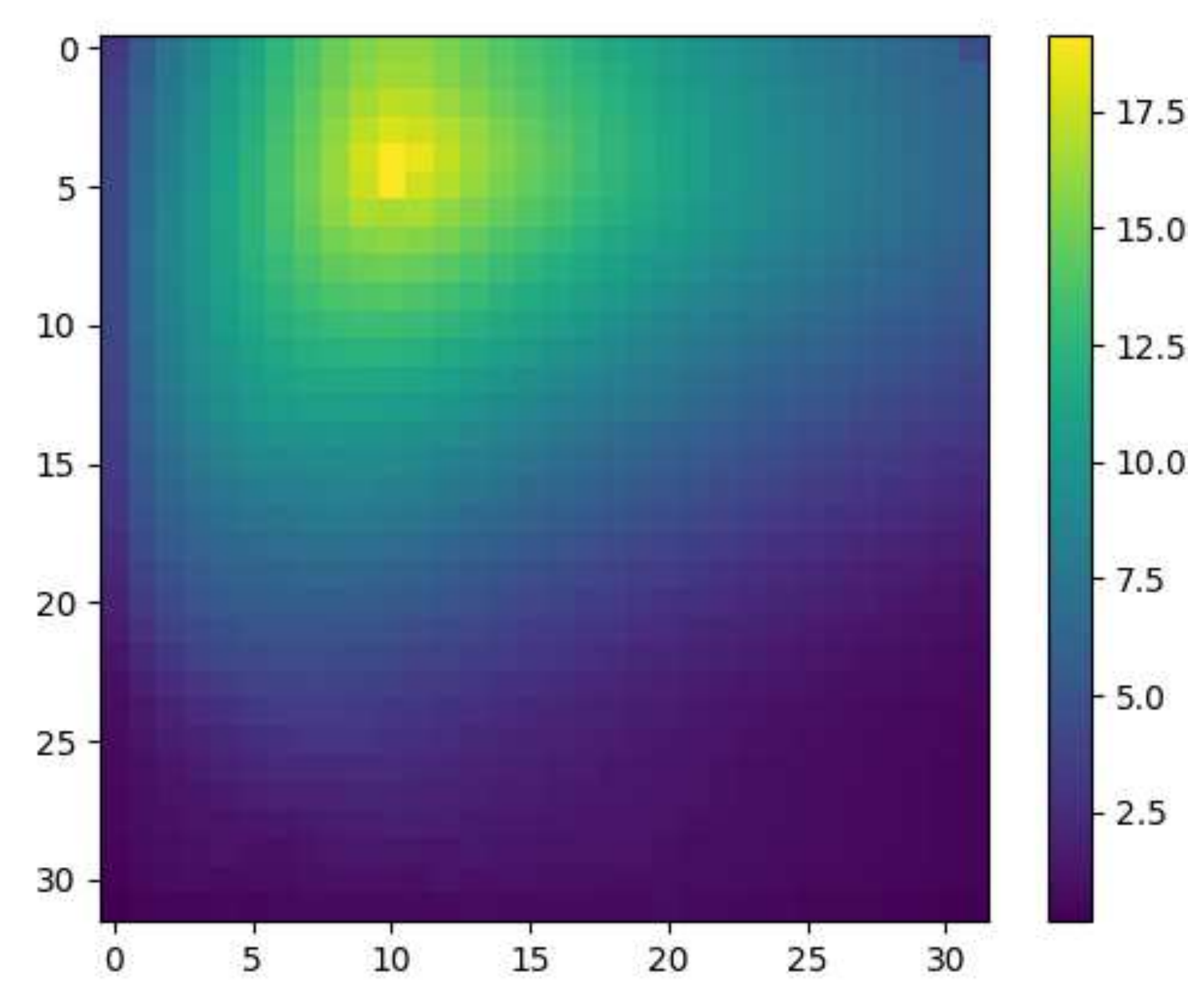}
\end{minipage}}
\caption{Value function estimated by different CNNs.}
\label{Fig.5.2}
\end{figure}

\subsubsection{Model Ablation Analysis of DB-CNN}
To evaluate whether DB-CNN could keep its performance after ablating some of its components, we compare DB-CNN with ResNet and DCNN, since ResNet ablates branch one of DB-CNN and DCNN replaces the residual layers on ResNet further. According to the results in TABLE~\ref{tab.4.1}, both ResNet and DCNN perform poor on planetary global path planning tasks. Specifically, without branch one, DB-CNN will lose its path planning accuracy on testing data. Moreover, without residual layers, training DB-CNN will be difficult, making it almost unable to plan path from original planetary images. Therefore, it can be concluded that the double branch structure of DB-CNN indeed contributes to its final performance on global path planning, and the residual layers can enhance the training efficieny of DB-CNN.

Furthermore, to explain why DB-CNN works well, we visualize the value function estimation results of DB-CNN, VIN and ResNet (we ignore DCNN due to its poor performace). Since the final layer of these architectures will output the estimated Q value ($\hat Q(s,a)$) for each input and rover's moving direction for next step, the state value function $\hat V(s,a)$ can be derived as Eq.~(\ref{eq2.5}) and Eq.~(\ref{eq2.6}), which is illustrated in Fig.~\ref{Fig.5.2}. It can be seen that the state value functions estimated by DB-CNN are more in coincidence with the original Martian orbital images compared with VIN and ResNet. It is clear that risky areas are darker (smaller value) and the lighter locations (larger value) are around target points in state value function estimated by DB-CNN. By contrast, ResNet without global deep features cannot estimate the value function as precisely as DB-CNN. VIN also fails to recognize risky areas of Martian images evidently. Since the paths for planetary rover planned by these architectures follows the locations with higher value according to Eq.~(\ref{eq2.5}) and Eq.~(\ref{eq2.6}), the accuracy and successful rate of global path planning are determined by the precision of value function estimation. Therefore, from Fig.~\ref{Fig.5.2}, we can find that DB-CNN indeed works better of planetary path planning tasks than other baseline architectures. 

\section{Conclusions}
In this paper, we first propose a novel DCNN architecture with double branches---DB-CNN to path path for planetary rovers directly from orbital images, which requires no prior knowledge about the planetary orbital images. Then, we present the complete global path planning algorithm based on DB-CNN. Moreover, through comparison experiments on two global path planning datasets, we demonstrate that DB-CNN achieves higher precision and efficiency on global path planning tasks compared with the existing best architecture---VIN. Finally, we analyze why DB-CNN works well through model ablation analysis and visualization analysis. In future research, more effective deep neural network architecture will be explored and the robustness of the architecture will be researched further.
\section{Acknowledgement}
This work was supported by the National Key Research and Development Program of China under Grant 2018YFB1003700, the Beijing Natural Science Foundation under Grant 4161001, the National Natural Science Foundation Projects of International Cooperation and Exchanges under Grant 61720106010, and by the Foundation for Innovative Research Groups of the National Natural Science Foundation of China under Grant 61621063.

\end{document}